\documentclass[pdflatex,sn-mathphys,table]{sn-jnl}

\jyear{2022}%

\theoremstyle{thmstyleone}%
\theoremstyle{thmstyletwo}%

\theoremstyle{thmstylethree}%
\usepackage{cleveref}
\usepackage{caption}
\usepackage{subcaption}
\usepackage{xcolor}

\begin{document}

\title[A novel DL approach for one-step Conformal Prediction approximation]{A novel Deep Learning approach for one-step Conformal Prediction approximation}

\author*[1]{\fnm{Julia A.} \sur{Meister}}\email{J.Meister@brighton.ac.uk}

\author[1]{\fnm{Khuong An} \sur{Nguyen}}\email{K.A.Nguyen@brighton.ac.uk}

\author[2]{\fnm{Stelios} \sur{Kapetanakis}}\email{Stelios@distributedanalytics.co.uk}

\author[3]{\fnm{Zhiyuan} \sur{Luo}}\email{Zhiyuan.Luo@rhul.ac.uk}

\affil*[1]{\orgdiv{Computing and Maths Division}, \orgname{University of Brighton}, \orgaddress{\city{Brighton}, \postcode{BN2 4GJ}, \state{East Sussex}, \country{United Kingdom}}}

\affil[2]{\orgname{Distributed Analytics Solutions}, \orgaddress{\street{17 Fawe Street}, \city{London}, \postcode{E14 6FD}, \country{United Kingdom}}}

\affil[3]{\orgdiv{Department of Computer Science}, \orgname{Royal Holloway University of London}, \orgaddress{\city{Egham}, \postcode{TW20 0EX}, \state{Surrey}, \country{United Kingdom}}}


\abstract{Deep Learning predictions with measurable confidence are increasingly desirable for real-world problems, especially in high-risk settings. The Conformal Prediction (CP) framework is a versatile solution that guarantees a maximum error rate given minimal constraints~\cite{shafer2008tutorial}. In this paper, we propose a novel conformal loss function that approximates the traditionally two-step CP approach in a single step. By evaluating and penalising deviations from the stringent expected CP output distribution, a Deep Learning model may learn the direct relationship between the input data and the conformal p-values. We carry out a comprehensive empirical evaluation to show our novel loss function's competitiveness for seven binary and multi-class prediction tasks on five benchmark datasets. On the same datasets, our approach achieves significant training time reductions up to 86\% compared to Aggregated Conformal Prediction (ACP, \cite{norinder2016conformal}), while maintaining comparable approximate validity and predictive efficiency.}





\maketitle

\section{Introduction}
    \emph{How confident are Deep Learning predictions?} In most cases, we assume that a model performs as well on new data as it has on average in the past \cite{rechkemmer2022confidence,zhang2021confidence,meister2022audio,yin2019understanding}. The Conformal Prediction (CP) framework presents an enticing alternative, providing per-prediction confidence based on statistical hypothesis testing \cite{vovk2005algorithmic}. Given a user-specified significance level, standard CP guarantees a corresponding maximum error rate with more relaxed assumptions than commonly assumed in Deep Learning \cite{shafer2008tutorial}. However, guaranteed absolute validity comes with computational efficiency drawbacks, making CP in its original form unrealistic for large-scale datasets \cite{riquelme2019coreset}.
    
    Several proposals have been made in recent years that successfully address CP challenges, although they tend to introduce a trade-off with other model characteristics \cite{linusson2017calibration}. For example, inductive variants remove the need for repeated leave-one-out training \cite{papadopoulos2007conformal}. Aggregated CP (ACP) improves predictive efficiency (i.e.,\ the precision of the predictions), at the cost of training multiple ensemble models and losing absolute validity \cite{norinder2016conformal}. However, to the best of our knowledge, such approaches maintain the underlying two-step CP algorithm: calculating data strangeness with an intermediate non-conformity measure, and subsequently transforming scores into p-values.
    
    Distribution approximation provides an interesting avenue to address challenges related to the two-step nature of the Conformal Prediction algorithm. In this paper, we propose a one-step CP approximation approach with a novel, model-agnostic conformal loss function. By evaluating the deviation of the model output from the expected output distribution, we may circumvent the algorithmic constraints of the CP framework. Deep Learning is especially promising for this approach because the model's versatility and potential to model complex data relationships is well-established \cite{lecun2015deep,maskara2019advantages,khatri2019prediction}.

    \subsection{Contributions}
        To extend CP usability on real-world datasets, we make the following contributions:
        \begin{itemize}
            \item \emph{We propose a unique loss function that approximates traditional two-step CP in only one step} (\Cref{sec:CP,sec:proposedLoss}). \\           
            By enabling a Deep Learning (DL) neural network to learn the direct relationship between input data and the conformal p-values, we may skip the intermediate non-conformity scores altogether.

            \item \emph{We carry out a rigorous and comprehensive evaluation of our proposed method for 7 binary and multi-class classification tasks on 5 benchmark datasets} (\Cref{sec:results}). \\
            Empirical analysis of DL models trained with our loss function confirms competitiveness with Aggregated Conformal Prediction (ACP) in terms of approximate validity and predictive efficiency. However, our model is significantly more computationally efficient than ACP.
            
            \item \emph{Finally, our one-step approach to CP approximation introduces new opportunities for CP improvement} (\Cref{sec:discussion}). \\            
            By modelling the expected output distribution directly, our novel loss function approach circumvents current CP algorithmic constraints. This is of particular research interest, as CP performance optimisation is traditionally a challenging task due to the interactions between the underlying model and the non-conformity measure \cite{cherubin2021exact}.
        \end{itemize}
        
    The code and results described in this article are available in a GitHub repository\footnote{\url{https://github.com/juliameister/dl-confident-loss-function}}.

\section{Conformal Prediction}\label{sec:CP}    
    This section outlines the CP background, which the article builds on.
    
    \subsection{Background}\label{sec:cpBackground}
        The CP framework answers a ubiquitous question in Machine Learning: \emph{How confident are we that a model's prediction is correct?} In this article, the focus lies on a computationally efficient variant, Inductive Conformal Prediction (ICP) \cite{vovk2012conditional}. Interested readers may refer to \cite{vovk2005algorithmic} for a detailed description and context of the original transductive approach (TCP).
        
        Under minimal data assumptions, CP provides guaranteed confidence for individual predictions, based on statistical hypothesis testing \citep{shafer2008tutorial}. Specifically, it guarantees that a predictor makes errors only up to a maximum user-specified error rate $\epsilon$  if the input data is exchangeable \cite{gupta2022nested}. CP achieves this by outputting prediction sets with all plausible labels for classification problems (interested readers may refer to \cite{johansson2014regression} for regression). An error occurs when the true label $y_i$ is not included in the CP prediction set $\Gamma^{1-\epsilon}_i$. By the law of large numbers, the probability of an error occurring approaches the upper limit $\epsilon$ as the number of predictions grows (\Cref{eq:validity}), subject to statistical fluctuations \cite{fisch2021few}.
            \begin{align}
                Pr(y_i\not\in\Gamma^{1-\epsilon}_i) \leq \epsilon \label{eq:validity}
            \end{align}
            
        In short, the CP framework acts as a wrapper for any prediction model, also called the underlying model \cite{lofstrom2013effective}. Predictions are transformed into p-values $p_i^{y^*}$ that describe the likelihood of sample $x_i$'s extension with each possible label $y_i^* \in Y$, denoted $z_i = (x_i, y_i^*)$. The computationally efficient inductive CP variant (ICP) requires three sub-datasets. Let the proper training set be $(x_1, ..., x_n)$, the calibration set $(x_{n+1}, ..., x_m)$, and the test set $(x_{m+1}, ..., x_l)$. For each sample extension $z_i$, the ICP model first evaluates how strange it is compared to the known training data \cite{norinder2017binary}. The extension's `strangeness', or non-conformity, is measured with a score $\alpha_i^{y^*}$ as a function of the extended sample $z_i$ and its true label $y_i$. An example of a straightforward but versatile non-conformity measure (NCM) is the Margin Error function $A_{ME}$, given in \Cref{eq:acpNCM}. A non-conforming example has a low probability estimate for the true label $y_i$, and/or a high probability estimate for the false label $y_i^* \neq y_i$ \cite{johansson2017model}.
            \begin{gather}
                \alpha_i^{y^*} = A(z_i, y_i) \\
                A_{ME}(z_i, y_i) = 0.5 - \frac{\max_{y_i \neq y_i^*} Pr(y_i \text{\textbar} x_i) - Pr(y_i^* \text{\textbar} x_i)}{2} \label{eq:acpNCM}
            \end{gather}
    
        For ICP, the model is fit to the data just once \cite{johansson2013conformal}. This includes training the underlying algorithm with the proper training set $(x_1, ..., x_n)$, and calculating the NCM scores $\alpha_{n+1}^{y_{n+1}}, ..., \alpha_{m}^{y_m}$ for the calibration set $(x_{n+1}, ..., x_m)$. Given a test score $\alpha_{i}^{y_{i}^*}$, we calculate the p-values $p_{i}^{y_{i}^{*}}$ for each $z_i \in (z_{m+1}, ..., z_l)$, as shown in \Cref{eq:pvalues}. The numerator is also known as a `rank'. The larger the rank, the larger the p-value is, and the more conforming a sample extension $z_i$ is, compared to the training and calibration sets \cite{vovk2013transductive}. Furthermore, \Cref{eq:pvalues} guarantees that all p-values corresponding to the true labels $y_i$ are uniformly distributed in $\mathcal{U}_{[0, 1]}$ \cite{angelopoulos2020uncertainty}.
            \begin{align}
                p_i^{y^{*}} &= \frac{\text{\textbar}\{j=n+1,...,m:\alpha^{y^*}_j \leq \alpha^y_{i}\}\text{\textbar}+1}{\text{\textbar}\{j=n+1, ..., m: y^*_j\}\text{\textbar}+1} \label{eq:pvalues}
            \end{align}
        
        Finally, given a test sample's p-values $p_{m+1}^{y_{i}^{*}}, y_i^* \in Y$, its CP prediction set $\Gamma_i$ is obtained with \Cref{eq:predictionSets}. The optimal and most precise CP classification output is a prediction set with exactly one class label.
            \begin{gather}
                \Gamma_i^{1-\epsilon} = \{y^* \in Y \text{\textbar } p_i^{y^*} > \epsilon\} \label{eq:predictionSets} \\
                \text{\textbar}\Gamma_{optimal}^{1-\epsilon}\text{\textbar} = 1
            \end{gather}
            
        Note that, since the error rate and its inverse accuracy are automatically guaranteed by the validity characteristic, CP performance evaluation is traditionally based on predictive efficiency, i.e.,\ prediction set size \cite{krstajic2021critical}. \Cref{sec:evalMetrics} discusses metrics that may be used to evaluate CP performance.

    \subsection{Related work in conformal validity approximation}\label{sec:relatedWork}   
        Optimising CP for both classification and regression tasks is an active research area \cite{sesia2021conformal,papadopoulos2011regression,linusson2021nonconformity} because the framework has promising applications in high-risk, confidence-sensitive settings. CP is versatile, and a small selection of use cases include pharmaceutical drug discovery \cite{eklund2015application}, respiratory health monitoring \cite{meister2020conformal,nguyen2018cover}, and financial risk prediction \cite{wisniewski2020application}.
        
        Since prediction accuracy is automatically guaranteed for traditional `full' or Transductive CP (TCP), the primary optimisation objective is to increase predictive efficiency by outputting more precise prediction ranges (\Cref{sec:cpBackground}). Unfortunately, optimisation is a difficult task. There are two related but sometimes conflicting steps which must be considered simultaneously: the calibration of the underlying predictive model and the non-conformity measure. While promising approaches have been proposed that take both steps into account \cite{cherubin2021exact,makili2013incremental,papadopoulos2009reliable}, they tend to be specific to certain underlying models and not generally applicable.
    
        Additionally, TCP has computational efficiency challenges built into the algorithm and is not scalable to large, real-world datasets \cite{riquelme2019coreset}. CP variants that address the inefficiency of the transductive leave-one-out retraining approach tend to accept a trade-off with other limitations \cite{linusson2017calibration}. For example, Inductive Conformal Prediction (ICP) trades predictive efficiency for increased computational efficiency~\cite{papadopoulos2007conformal}.
        
        In contrast, Aggregated Conformal Prediction (ACP) improves predictive efficiency by limiting prediction set sizes \cite{norinder2016conformal}. It is a generalisation of previously proposed CP variants (Cross-conformal and Bootstrap conformal predictors, introduced in \cite{vovk2015cross}), inspired by ML prediction ensembles to reduce variance. For each test sample $x_i$ and possible label $y^* \in Y$, the intermediate p-values $p_{i_n}^{y^*}$ generated by $N$ ensemble models are aggregated into one ACP p-value $\hat{p}_i^{y^*}$ (e.g., with \Cref{eq:acpPvalue}, \cite{carlsson2014aggregated}). The intermediate p-values are calculated as shown in \Cref{sec:cpBackground}.
        \begin{gather}
            \hat{p}_i^{y^*} = \frac{1}{N} \sum^N_{n=1} p_{i_n}^{y^*} \label{eq:acpPvalue}
        \end{gather}
        However, more precise ACP predictions come at the cost of losing automatic validity. While exact validity may be achieved with more stringent requirements or for some underlying models, approximate validity is observed empirically \cite{wilm2020skin}. In particular, ACP tends to be conservative for low significance levels and invalid for high significance levels, because p-values close to the mean tend to be more common \cite{linusson2017calibration}. There have been some effective proposals to recover validity \cite{solari2022multi,toccaceli2019combination,balasubramanian2015conformal}, but these approaches do not address the significantly increased training times for multiple ensemble models compared to one ICP model.
        
        In a recent and particularly relevant work, non-conformity scores were directly approximated by estimating their influence on the underlying model's loss with Influence Functions \citep{abad2022approximating}. The purpose was to make the algorithm scalable to large datasets. The method successfully achieves validity on par with TCP, while removing the need for a leave-one-out training procedure for every test point. As the number of training samples increased, the authors observed an increasingly diminishing approximation error. In empirical experiments, the error became negligible at 10,000 training samples.
        
        This confirms that validity and improved computational efficiency are possible by approximating individual steps of the TCP algorithm. In this article, we propose a novel conformal loss function that approximates the entire TCP two-step procedure by teaching a Deep Learning model the direct relationship between input data and conformal p-values.

\section{Approximating conformal p-values directly with Deep Learning}\label{sec:proposedLoss}
    Deep Learning (DL) is a highly versatile framework that automatically identifies and learns relationships between input data and output data, by extracting informative data representations \cite{lecun2015deep}. A neural network $h$ with weights $\theta$ is made up of one or more hidden and interlinked layers (\Cref{eq:nnFormula}). Each hidden neuron $H_i$ in a layer performs a simple linear transformation on the incoming data vector $X$ with weight vector $\theta_i$, offset by an absolute bias $B_i$, as shown in \Cref{eq:neuronFormula} \citep{ghalambaz2011hybrid}. An activation function (e.g., ReLU) may then be applied to model non-linear data relationships.
    \begin{gather}
        \hat{y} = h_{\theta} (X) \label{eq:nnFormula} \\
        H_i = B_i + \theta_{i,1} \cdot X_1 + \theta_{i,2} \cdot X_2 + ... + \theta_{i,j} \cdot X_j \label{eq:neuronFormula}
    \end{gather}
    Supervised training may be considered as an optimisation function with which the model attempts to minimise the deviation of its current output prediction $\hat{y}$ to the expected output label $y$ \cite{olive2018supervised}. For DL models trained with gradient descent, the difference between them is quantified via a loss function $f$ (\Cref{eq:loss}).
    \begin{align}
        C = f(\hat{y}, y) \label{eq:loss}
    \end{align}
    The loss $C$ is then back-propagated through the model. In \Cref{eq:backprop}, each weight $\theta_i$ is updated with $\Delta \theta_i$ in proportion to its negative gradient \citep{lillicrap2020backpropagation}, reducing the output's deviation from the true labels in the next iteration.
    \begin{align}
        \Delta \theta_{i} = - \frac{\partial C}{\partial \theta_{i}} \label{eq:backprop}
    \end{align}
    Once training is complete, a well-calibrated model will have optimised its learned weights $\theta$ so that \Cref{eq:nnFormula} accurately reflects the relationship between the input and its expected output label.
    
    DL's versatility and ability to represent complex data relationships make it a promising choice to model conformal p-values directly from the input, skipping the non-conformity scores (see \Cref{sec:CP}). We hypothesise that a function composition of the two standard CP transformation steps (from data input to non-conformity measures, and from non-conformity measures to p-values) could be approximated with a suitably designed and trained model. \emph{In other words, the right DL architecture and loss function may be able to model the complex relationship between the input and conformal p-value output directly.}
    
    DL architectures have been used in the past to predict probability distributions of the target variable, e.g., with Mixture Density Networks \cite{makansi2019overcoming,li2019generating,zhang2020improved}. However, to our knowledge, there is currently no model architecture (e.g.,\ activation function) or loss function that can guarantee a uniform model output distribution, which is required for true-class conformal p-values (\Cref{sec:cpBackground}). Therefore, we propose an architecture-agnostic conformal loss function which ensures that the model output approximates the uniform distribution $\mathcal{U}_{[0, 1]}$.

    The purpose of our novel loss function is to achieve approximate validity and high precision, while significantly reducing the high algorithmic and computational complexity of similar CP variants. For example, ACP uses an ensemble approach to improve predictive efficiency \cite{norinder2016conformal}. However, as the number of ensemble models grows, computational time increases significantly. Additionally, approximate validity becomes weaker because the distribution of aggregated p-values shifts away from $\mathcal{U}_{[0,1]}$ (e.g., to a unimodal Bates distribution with the aggregation procedure shown in \Cref{eq:acpPvalue}) \cite{linusson2017calibration}.
    
    In contrast, a single DL model with our proposed conformal loss function is competitive with ACP up to 10 ensemble models in terms of approximate validity and high predictive efficiency, while significantly reducing algorithmic complexity and training time. \Cref{sec:proposalRequirements,sec:proposedMethod} describe the background and components of our proposed method in detail, and \Cref{sec:results} provides a rigorous empirical evaluation of our results on the well-established MNIST datasets \cite{lecun1998gradient}.

    \subsection{Proposed method requirements and constraints}\label{sec:proposalRequirements}
        This article's contribution is to develop a unique conformal loss function for Deep Learning (DL). The novelty of our approach is that we simplify the two-step CP algorithm (\Cref{sec:CP}) to only one step. Our loss function approximates conformal p-values directly from the input data, skipping the intermediate non-conformity measure. Therefore, the neural network's output should follow the same distribution requirements as CP p-values:
        \begin{itemize}
            \item To \emph{interpret the neural network output as conformal p-values}, predictions should fall in the range (0, 1), and class outputs for one sample should be independent (\Cref{sec:cpBackground}).
            \item To \emph{ensure validity and consequently a guaranteed error rate}, approximated p-values of the true class should be uniformly distributed in $\mathcal{U}_{[0, 1]}$ \cite{linusson2017calibration}.
            \item To \emph{maintain high predictive efficiency}, false-class approximated p-values should be close to 0 \cite{shafer2008tutorial}.
        \end{itemize}
        Additionally, we must work within the constraints enforced by DL:
        \begin{itemize}
            \item The \emph{loss function should be differentiable} for effective backpropagation with gradient descent \cite{lecun2015deep}.
            \item The \emph{trade-off between model genralisability} (small batch size, \cite{kandel2020effect}) and \emph{output distribution evaluation precision} (large batch size, \cite{hao2019doubly}) should be considered.
        \end{itemize}

    \subsection{Proposed conformal approximation loss function and background}\label{sec:proposedMethod}
\begin{figure}[htbp]
            \centering
            \begin{subfigure}{0.32\textwidth}
                \centering
                \includegraphics[width=\textwidth]{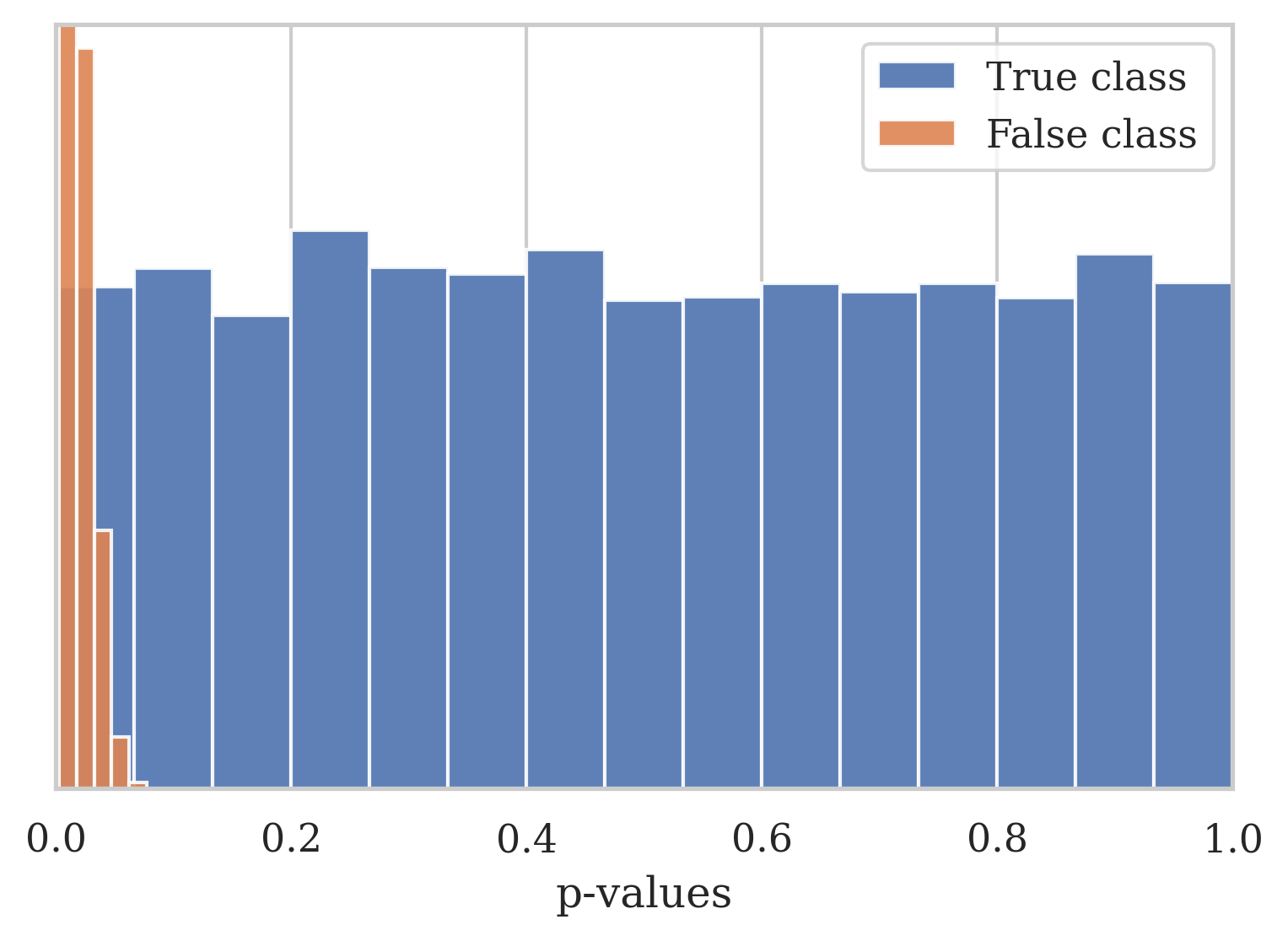}
                \caption{Well-calibrated CP.}
                \label{fig:cpWellCalibrated}
            \end{subfigure}
            \hfill
            \begin{subfigure}{0.32\textwidth}
                \centering
                \includegraphics[width=\textwidth]{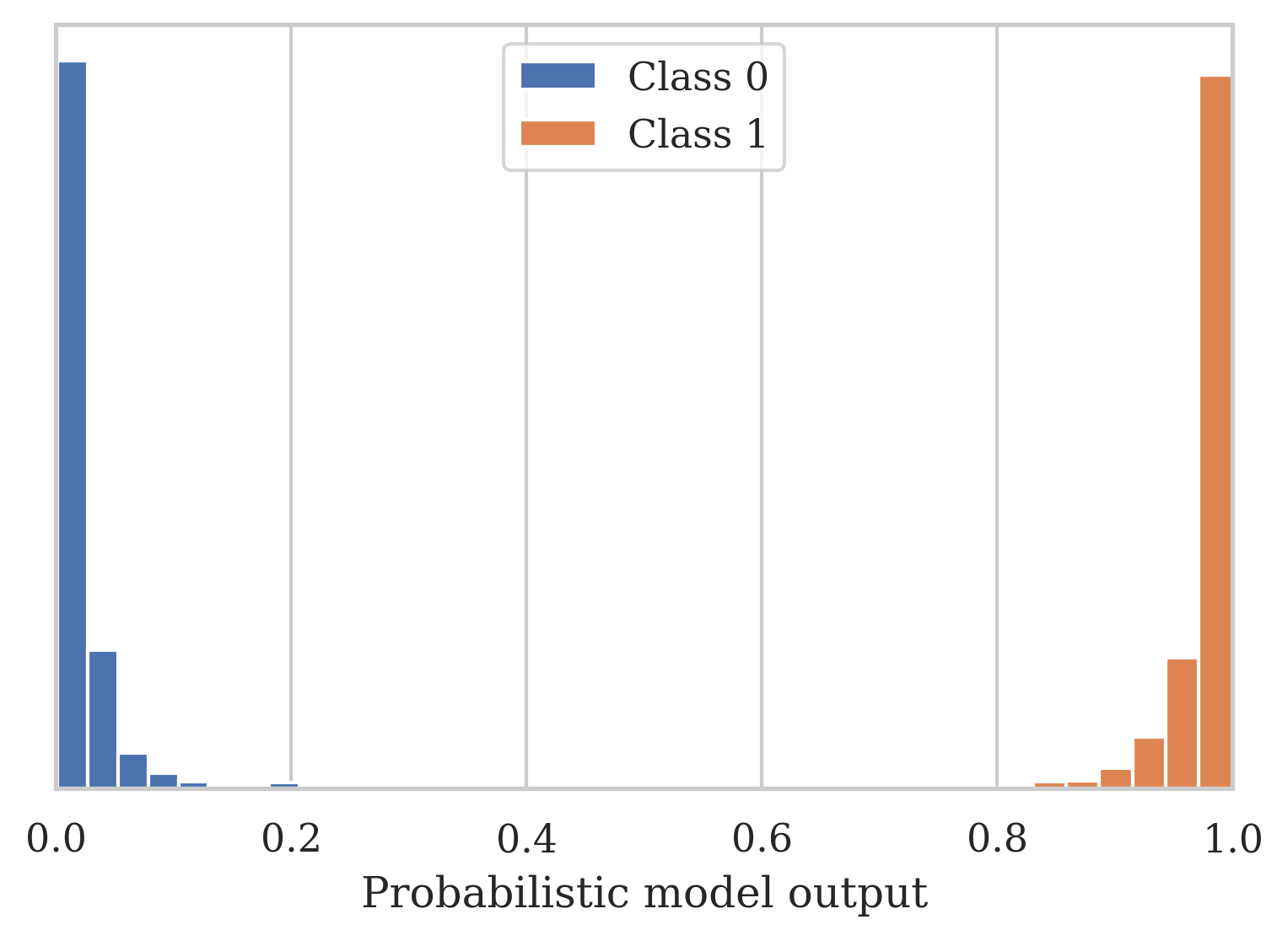}
                \caption{Well-calibrated NN.}
                \label{fig:nnWellCalibrated}
            \end{subfigure}
            \hfill
            \begin{subfigure}{0.32\textwidth}
                \centering
                \includegraphics[width=\textwidth]{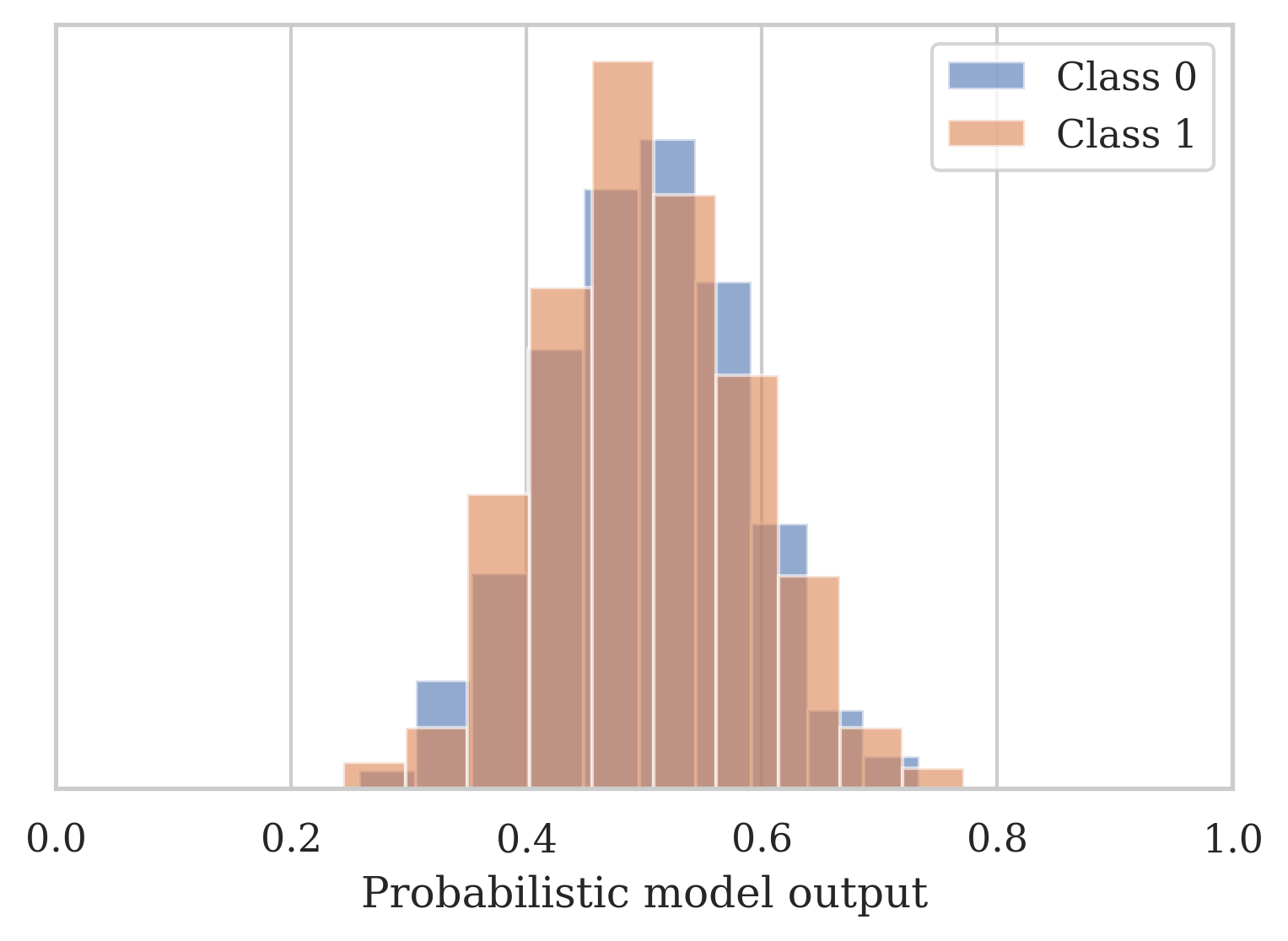}
                \caption{Underfit NN.}
                \label{fig:nnBadlyCalibrated}
            \end{subfigure}
            \caption{Characteristic Neural Network (NN) and Conformal Prediction (CP) output distributions of binary probabilistic classifiers.}\label{fig:calibration}
        \end{figure}

        To approximate conformal p-values directly from the input data, we propose a conformal loss function that is compatible with a probabilistic, binary classification neural network (architecture details in \Cref{sec:results}). Given the requirements of the CP framework (\Cref{sec:CP}), we know that well-calibrated CP classifiers are marked by a uniform distribution of the true-class p-values, and a distribution peaking at 0 with little variance for false-class p-values \cite{linusson2017calibration}, as visualised in \Cref{fig:cpWellCalibrated}. In contrast, typical well-calibrated neural network (NN) classifiers have a distinctive maximum distance between true-class (peak at 1) and false-class (peak at 0) output distributions, as shown in \Cref{fig:nnWellCalibrated}. The challenge for the conformal loss function is to accurately describe the CP output distribution, so that the neural network may emulate it, avoiding the normal-like distribution that tends to occur when under-fit models make average random guesses (\Cref{fig:nnBadlyCalibrated}). Therefore, we propose a loss function that is made up of two components, optimising the true-class and false-class target distributions, respectively:
        \begin{align}
            loss_{conformal} = loss_{false} + loss_{true} \label{eq:lossTotal}
        \end{align}
        
        Since approximated p-values of the false class should follow the same pattern as traditional NN false-class predictions, we calculate standard Binary Cross Entropy (BCE) to minimise the values towards 0. \Cref{eq:bce} defines BCE as a function of the model's predictions $\hat{y}$ and the true labels $y$ for $N$ training samples \citep{ho2019real}. For the purpose of the conformal loss function, we calculate BCE only for the false-class approximated p-value predictions $\hat{y}_{false}$ (\Cref{eq:pvaluesFalse,eq:lossFalse}).
        \begin{gather}
            BCE(y, \hat{y}) = - \frac{1}{N} \sum_{n=1}^{N} (y_n \cdot \log \hat{y}_n + (1-y_n) \cdot \log (1-\hat{y}_n)) \label{eq:bce} \\
            \hat{y}_{false} = [n=1, ..., N,  y_n \in \mathbf{Y}^{\mathbb{N}}: \hat{y}_{nc} \text{ s.t. } c \neq y_n] \label{eq:pvaluesFalse} \\
            loss_{false} = BCE(0, \hat{y}_{false}) \label{eq:lossFalse}
        \end{gather}

        Unlike traditional classification, the true labels $y \in \mathbf{Y}^\mathbb{N}$ are not the target model output in CP. Instead, true-class predictions $\hat{y}_{true}$ (\Cref{eq:pvaluesTrue}) should follow the distribution $\mathcal{U}_{[0, 1]}$ over $N$ samples (\Cref{sec:CP}). As a consequence, the loss component $loss_{true}$ should evaluate and quantify the distance of the model output distribution from the target uniform distribution, rather than the distance to a concrete label per sample. Note that the trade-off between large sample counts for distribution estimation \cite{jaki2019effects} and small batch sizes for optimised learning \cite{kandel2020effect} must be carefully considered.
        
        Even though function moments do not necessarily uniquely characterise a distribution, they may be a sufficient heuristic for a distribution distance evaluation \cite{malz2018approximating}. Lower order moments require fewer samples to estimate and are therefore suitable for the proposed true-class conformal loss component, shown in \Cref{eq:loss}. When $loss_{true}$ is minimised, the distribution of $\hat{y}_{true}$ approximates $\mathcal{U}_{[0, 1]}$. In \Cref{sec:results}, we empirically confirm that our one-step approximation approach is competitive with ACP in terms of uniformity and predictive efficiency, and significantly improves the computational efficiency.
        \begin{gather}
            \hat{y}_{true} = [n=1, ..., N, y_n \in \mathbf{Y}^{\mathbb{N}}: \hat{y}_{nc} \text{ s.t. } c=y_n] \label{eq:pvaluesTrue} \\
            loss_{true} = w_1 \cdot loss_{mean} + w_2 \cdot loss_{var} + w_3 \cdot loss_{l2} + w_4 \cdot loss_{huber} \label{eq:lossTrue}
        \end{gather}
        
        Including the \emph{mean} $\mu$ of true-class model outputs $\hat{y}_{true}$ ensures that the output distribution is centred around 0.5 (\Cref{eq:lossMean}). Because model training minimises the overall loss, we square and take the root, so that any deviation is represented by a positive value. A larger value indicates a greater distance between the model's output distribution and the expected mean of the uniform distribution $\mathbb{E}[\mathcal{U}_{[0, 1]}] = 0.5$.
        \begin{align}
            \mu(x) &= \frac{\sum_{n=1}^N x_n}{N} \label{eq:mean} \\
            loss_{mean} &= \sqrt{(\mu(\hat{y}_{true})-0.5)^2} \label{eq:lossMean}
        \end{align}
        
        Similarly, for the \emph{variance} $\sigma^2$, the $loss_{var}$ component regulates the dispersion of $\hat{y}_{true}$ so that it matches the expected value $\text{Var}[\mathcal{U}_{[0, 1]}] = \frac{1}{12}$.
        \begin{align}
            \sigma^2(x) &= \frac{\sum_{n=1}^N (x_n - \mu)^2}{N} \label{eq:var} \\
            loss_{var} &= \sqrt{\left( \sigma^2(\hat{y}_{true})-\frac{1}{12}\right)^2 } \label{eq:lossVar}
        \end{align}
        
        To measure uniformity while maintaining differentiability (i.e.,\ without ranking or sorting), we leverage the work in \cite{batu2017generalized}. The authors suggested that for an unknown distribution, measuring sample collisions with moments (such as $l_p$-norms) was a successful approximation for uniformity testing.

        We choose the \emph{$l_2$-norm} as our heuristic, since it is fully differentiable and sensitive to outliers, i.e.,\ it will penalise large differences more than smaller ones \citep{meyer2021alternative}. In \Cref{eq:lossL2}, $\hat{y}_{true}$ is normalised to ensure that minimising the loss component increases the uniformity of the model output distribution. Note that $loss_{l2}$ does not necessarily measure uniformity in the range $(0,1)$, only the distribution probability's smoothness between the input vector's extreme elements. The distribution's mean, variance, and the next loss component $loss_{huber}$ counteract this drawback by encouraging values throughout the entire range.
        \begin{align}
            l_2(x) = \sqrt{\sum_{n=1}^N (x_n)^2} \label{eq:l2} \\
            loss_{l_2} = l_2\left(\frac{\hat{y}_{true}}{\Sigma \hat{y}_{true}}\right) \label{eq:lossL2}
        \end{align}
        
        Traditionally, \emph{Huber loss} (\Cref{eq:huber}) is used to narrow the prediction region to the true label. In \Cref{eq:lossHuber}, we negate the value instead to disincentivise the trend to normal-like distributions on the average output $\delta$ of under-fit models. The parameter $\alpha \in \mathbb{R}^+$ regulates the threshold for the transition between quadratic and linear \citep{meyer2021alternative}. Since the target distribution for $\hat{y}_{true}$ is $\mathcal{U}_{[0, 1]}$, we expect the over-represented prediction to be $\delta \approx 0.5$ (see \Cref{sec:results}). As with previous loss components, Huber loss is fully differentiable and robust to outliers.
        \begin{align}
            Huber_\alpha(x) =
                \begin{cases}
                    \frac{1}{2} x^2, & \text{\textbar}x\text{\textbar} \leq \alpha \\
                    \alpha(\text{\textbar}x\text{\textbar}-\frac{1}{2} \alpha), & \text{\textbar}x\text{\textbar} > \alpha \\
                \end{cases} \label{eq:huber} \\
            loss_{huber} = - Huber_{\alpha}(\delta, \hat{y}_{true}) \label{eq:lossHuber}
        \end{align}
        
        \Cref{algo:loss} illustrates how the described loss components come together to produce our novel conformal loss function (also shown in \Cref{eq:loss}). The function is fully differentiable and, therefore, may be used in combination with any backpropagation model. All individual evaluation metrics are inbuilt into \verb|tensorflow|, meaning that our proposed loss is compatible with any \verb|tensorflow| model. \Cref{algo:modelTraining} shows the \verb|Python 3.9| implementation as a \verb|tensorflow v2.4| custom loss function.
        
        \begin{algorithm}
            \caption{Proposed conformal loss function components. By evaluating the deviation from the expected Conformal Prediction output distributions (\Cref{sec:cpBackground}), the neural network model may approximate conformal p-values directly from the input.}\label{algo:loss}
            \begin{algorithmic}[1]
                \Require $\hat{y} \in (0, 1), \text{ } y \in \mathbb{N}, \text{ } \theta_{huber}={(\alpha, \delta}), \text{ } \theta = (\theta_0, \theta_1, ..., \theta_5)$
                \State $\hat{y}_{false} \Leftarrow [n=1, ..., N : \hat{y}_{nc} \text{ s.t. } c \neq y_n]$
                \State $\hat{y}_{true} \Leftarrow [n=1, ..., N : \hat{y}_{nc} \text{ s.t. } c=y_n]$
                \State $loss_{false} = BCE(zerosLike(\hat{y}_{false}), \hat{y}_{false})$
                \State $loss_{mean} \Leftarrow \sqrt{(mean(\hat{y}_{true})-0.5)^2}$
                \State $loss_{var} \Leftarrow \sqrt{(var(\hat{y}_{true})-1/12)^2}$
                \State $loss_{l2} \Leftarrow l2(y_{true}/\Sigma \hat{y}_{true})$
                \State $loss_{huber} \Leftarrow - Huber_{\alpha}(\delta, \hat{y}_{true})$
                \State $loss_{true} \Leftarrow \theta_2 \cdot loss_{mean} + \theta_3 \cdot loss_{var} + \theta_4 \cdot loss_{l2} + \theta_5 \cdot loss_{huber}$
                \State $loss_{conformal} \Leftarrow \theta_0 \cdot loss_{false} + \theta_1 \cdot loss_{true}$
            \end{algorithmic}
        \end{algorithm}
  
\begin{minipage}{\hsize}%
\lstset{frame=single,framexleftmargin=-1pt,framexrightmargin=-17pt,framesep=11pt,linewidth=0.98\textwidth,language=Python,breaklines=true}
\captionof{algorithm}{Conformal loss implementation of \Cref{algo:loss} in Python 3.9. All classes and methods were imported from TensorFlow v2.4.}\label{algo:modelTraining}
\begin{lstlisting}
def conformal_loss(y_true, y_pred, alpha, delta, ws):
    # label formatting
    y_onehot = one_hot(reshape(y_true, [-1], depth=n_class)
    mask_true = cast(y_onehot, bool)
    mask_false = logical_not(mask_true)
    y_true = reshape(boolean_mask(y_pred, mask_true, axis=0), [-1, 1]) 
    y_false = reshape(boolean_mask(y_pred, mask_false, axis=0), [-1, n_class-1])
    
    # calculating loss components
    loss_false = BinaryCrossentropy(
        zeros_like(y_false), y_false)
    loss_l2 = l2_loss(y_true/reduce_sum(y_true))
    loss_mean = sqrt(square(reduce_mean(y_true)-0.5))
    loss_var = sqrt(square(reduce_variance(y_true)-1/12))
    loss_huber = -Huber(alpha)(fill(shape(y_true), delta), y_true)
    loss_true = w_l2 * loss_l2 + w_h * loss_huber +
        w_m * loss_mean + w_v * loss_var
    
    return w_t * loss_true + w_f * loss_false
\end{lstlisting}
\end{minipage}

\section{Empirical results}\label{sec:results}
    This section comprehensively evaluates the empirical performance of our proposed one-step conformal loss function (\Cref{sec:proposedMethod}). In particular, we compare the validity, predictive, and computational efficiency to the well-established ACP approximation technique (\Cref{sec:cpBackground}) for both binary and multi-class classification on 5 benchmark datasets.
    
    \subsection{Research questions}\label{sec:researchQuestions}
        The following research questions informed and structured the performance evaluation of our novel conformal loss function:
        \begin{itemize}
            \item Can we directly model the relationship between the input data and the conformal p-values, skipping the intermediate non-conformity measure?
            \item Is our proposed approach competitive with the established ACP approximation in terms of validity and predictive efficiency?
            \item What benefits may approximating CP with a single-step approach have, compared to the traditional two-step calculations?
        \end{itemize}

    \subsection{Datasets}\label{sec:dataset}
        We rigorously evaluate our proposed method for 7 classification tasks on 5 benchmark datasets. Predetermined train-test splits that were stored in the datasets were maintained. All other datasets are split into 67\% training and 33\% test samples, stratified by class. To reduce spurious noise in the results, the same split was used for all test runs. Sample counts are provided in \Cref{tab:sampleStats}.

        \begin{table}[!h]
            \centering\setlength{\tabcolsep}{2.3pt}
            \caption{Sample counts for the 7 classification tasks, using 5 benchmark datasets. MNIST and USPS are listed for both binary and multi-class classification.}\label{tab:sampleStats}
            \begin{tabular}{>{}l>{}c>{}c>{}c>{}c>{}c>{}c>{}c}
                \toprule
                \textbf{Samples} & \multicolumn{1}{c}{\textbf{MNIST2}} & \multicolumn{1}{c}{\textbf{MNIST10}} & \multicolumn{1}{c}{\textbf{USPS2}} & \multicolumn{1}{c}{\textbf{USPS10}} & \multicolumn{1}{c}{\textbf{WINE}} & \multicolumn{1}{c}{\textbf{BANK}} & \multicolumn{1}{c}{\textbf{MSHRM}} \\
                \specialrule{.4pt}{2pt}{0pt}\midrule
                Train & 12,665 & 60,000 & 2,199 & 7,291 & 4,352 & 27,595 & 40,916 \\
                Test & 2,115 & 10,000 & 623 & 2,007 & 2,145 & 13,593 & 20,153 \\
                \midrule
                $\sum$ & 14,780 & 70,000 & 2,822 & 9,298 & 6,497 & 41,188 & 61,069 \\
                \specialrule{.4pt}{2pt}{0pt}\bottomrule
            \end{tabular}
        \end{table}

        \subsubsection{MNIST dataset}\label{sec:mnist}
\begin{figure}[!h]
                \centering
                \begin{subfigure}{0.48\textwidth}
                    \centering
                    \includegraphics[width=\textwidth]{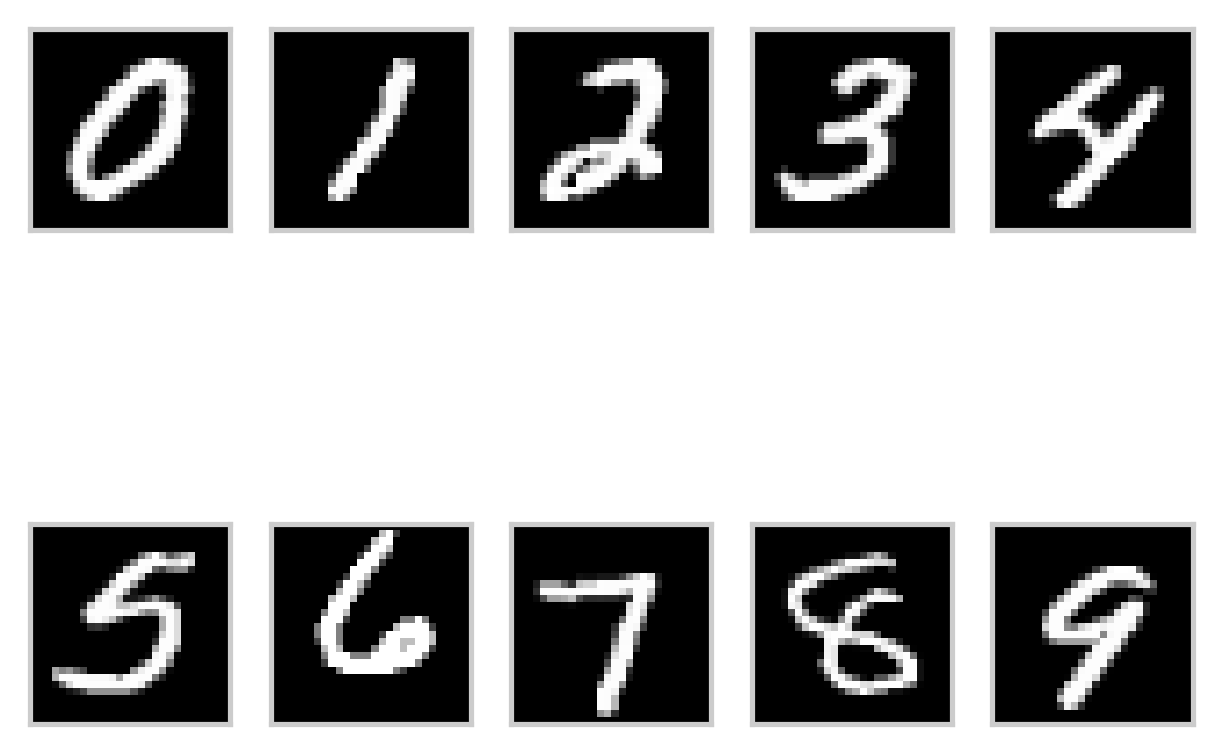}
                    \caption{Samples.}
                    \label{fig:mnistSamples}
                \end{subfigure}
                \hfill
                \begin{subfigure}{0.48\textwidth}
                    \centering
                    \includegraphics[width=\textwidth]{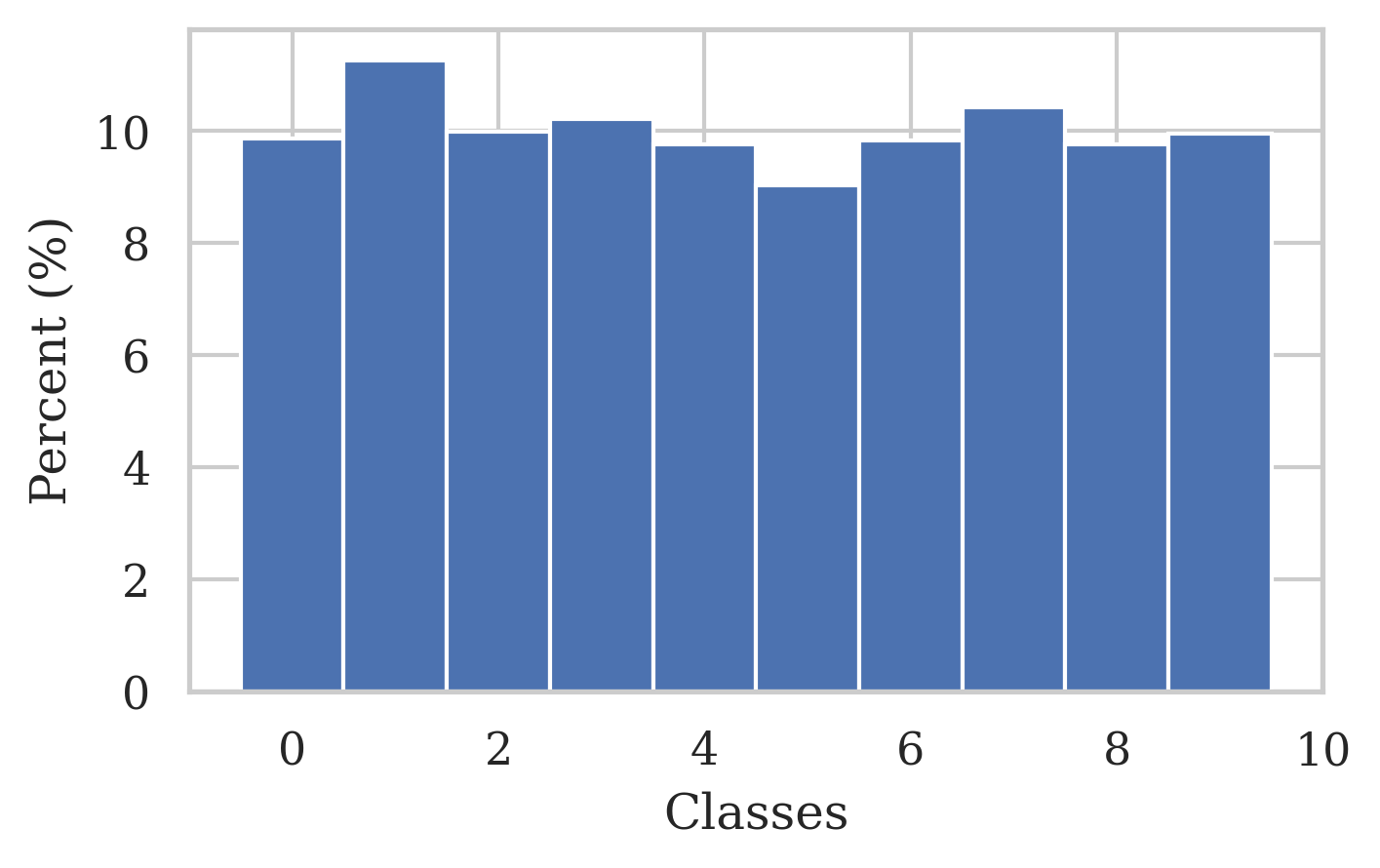}
                    \caption{Class distribution.}
                    \label{fig:mnistClasses}
                \end{subfigure}
                \caption{The MNIST dataset.}
            \end{figure}

            The MNIST dataset is an extensive collection of 70,000 images, each with a handwritten digit between 0--9~\cite{lecun1998gradient}. The greyscale images are formatted as 28x28 binary vectors (see \Cref{fig:mnistSamples}), flattened to (781 x 1) since we work with a linear feedforward network (see \Cref{sec:nnArchitecture}). We evaluate our proposed method for both binary classification (images of 0 and 1, referred to as \emph{MNIST2}) and multi-class classification with all 10 digits (referred to as \emph{MNIST10}). The classes are roughly balanced.

        \subsubsection{USPS dataset}
\begin{figure}[!h]
                \centering
                \begin{subfigure}{0.48\textwidth}
                    \centering
                    \includegraphics[width=\textwidth]{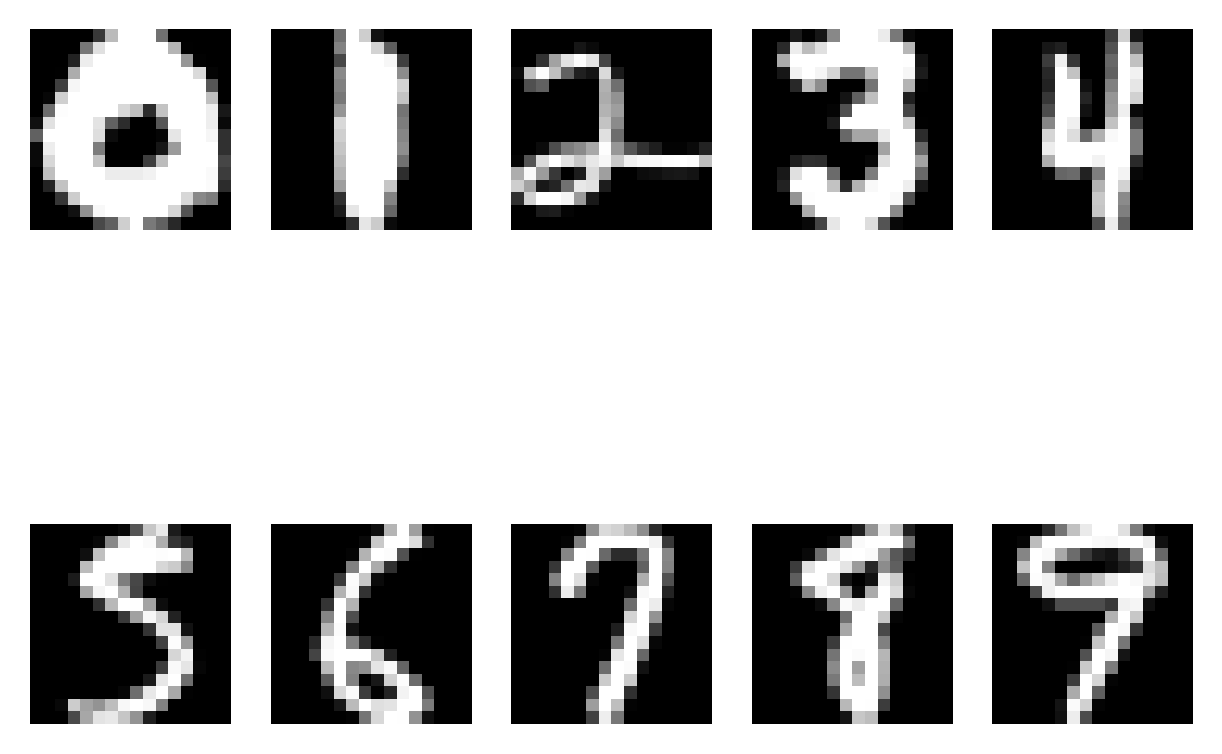}
                    \caption{Samples.}
                    \label{fig:uspsSamples}
                \end{subfigure}
                \hfill
                \begin{subfigure}{0.48\textwidth}
                    \centering
                    \includegraphics[width=\textwidth]{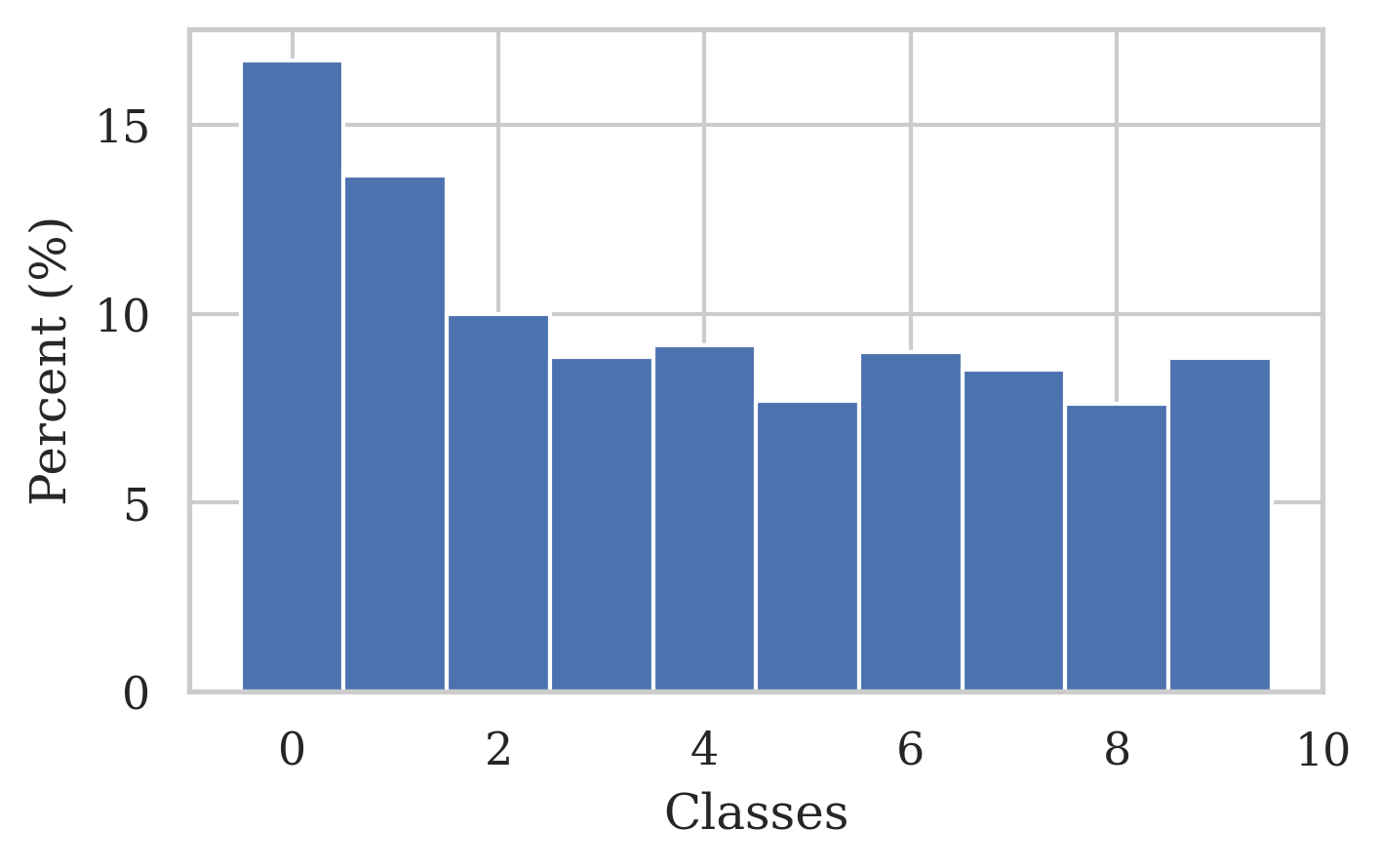}
                    \caption{Class distribution.}
                    \label{fig:uspsClasses}
                \end{subfigure}
                \caption{The USPS dataset.}\label{fig:uspsStats}
            \end{figure}

            The USPS dataset is a collection of handwritten digits in greyscale, collected from envelopes by the US postal service~\cite{hull1994database}. The images come in (16x16) images that are flatted to (256 x 1) feature vectors. Similarly to the MNIST dataset in \Cref{sec:mnist}, we consider two classification tasks: binary (USPS2) and multi-class (USPS10). Examples and the class distribution are visualised in \Cref{fig:uspsStats}.

        \subsubsection{Wine dataset}
\begin{figure}[!h]
                \centering
                \includegraphics[width=0.48\textwidth]{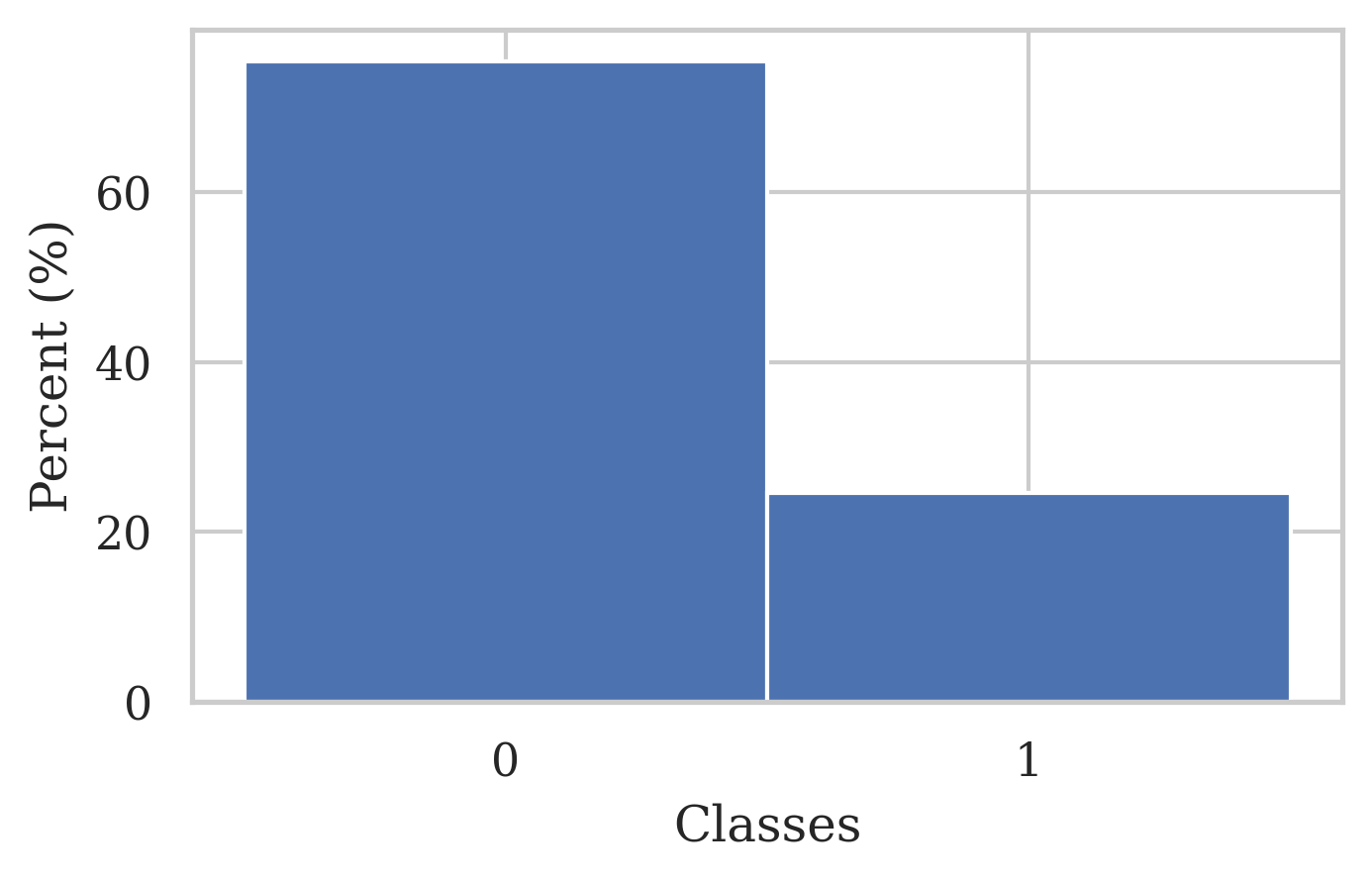}
                \caption{WINE dataset class distribution.}\label{fig:wineClasses}
            \end{figure}

            We use the Wine dataset to distinguish between red (class 1) and white (class 0) wines. We use 11 of the 12 features: Fixed acidity, volatile acidity, citric acid, residual sugar, chlorides, free sulphur dioxide, total sulphur dioxide, density, pH, sulphates, and alcohol. Wine quality is excluded because it is heavily imbalanced between the two classes. Readers are referred to~\cite{CorCer09} for detailed information about the features and \Cref{fig:wineClasses} for the class distribution.

        \subsubsection{Bank marketing dataset}
\begin{figure}[!h]
                \centering
                \includegraphics[width=0.48\textwidth]{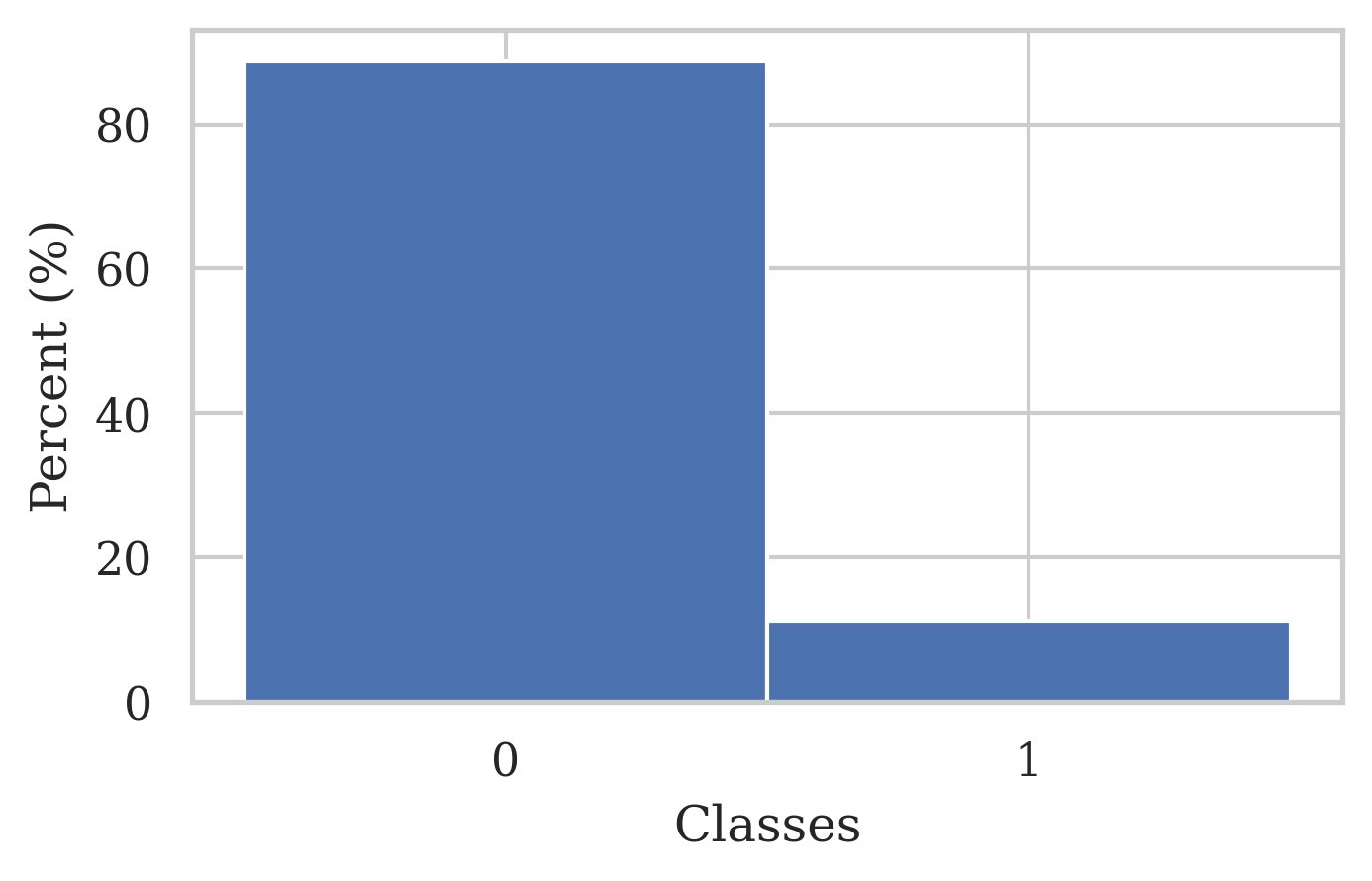}
                \caption{BANK dataset class distribution.}\label{fig:bankClasses}
            \end{figure}

            The Bank marketing dataset contains information about the outcomes of a mobile marketing campaign run by a Portuguese banking institution from 2008--2010. The classification task is to predict the success of the marketing based on 20 variables related to the client's lifestyle and when they were contacted. See~\cite{moro2014data} for more details. The class distribution is given in \Cref{fig:bankClasses}.
    
        \subsubsection{Mushroom dataset}
\begin{figure}[!h]
                \centering
                \includegraphics[width=0.48\textwidth]{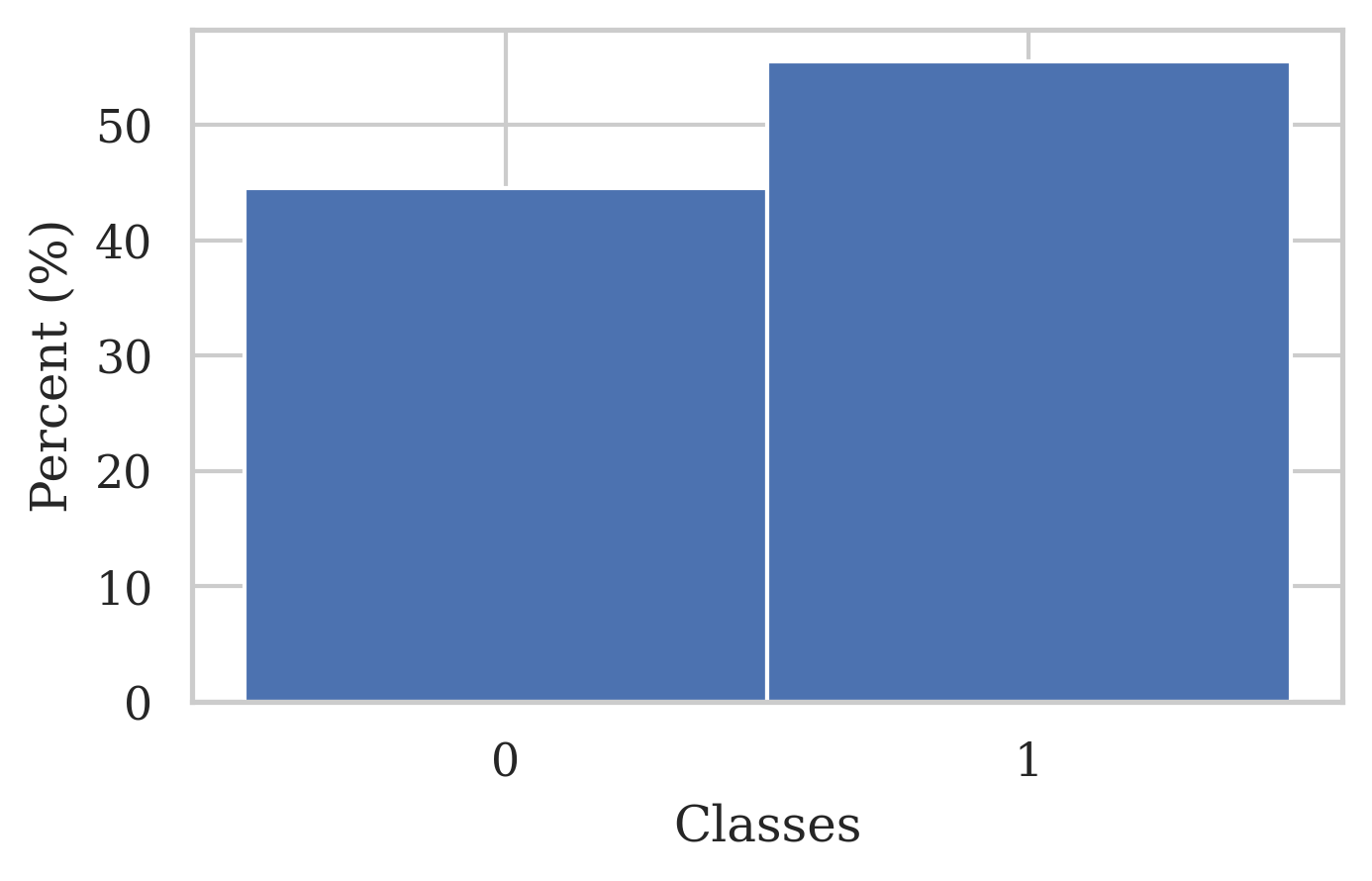}
                \caption{MSHRM dataset class distribution.}\label{fig:muschroomClasses}
            \end{figure}

            The Mushroom dataset~\cite{schlimmer1981mushroom} contains 22 characteristic features of mushroom species and whether they are edible (class 1) or poisonous (class 0) to humans. Categorical variables were replaced by dummy variables (e.g., stem colour). The distribution of classes is visualised in \Cref{fig:muschroomClasses}.
            
    \subsection{Evaluation metrics}\label{sec:evalMetrics}
        CP-related methods require unique metrics for performance evaluation since validity is guaranteed, unlike traditional Machine Learning models (\Cref{sec:CP}). Instead, we employ an intuitive performance measure related to the size of the CP models prediction sets \citep{ashby2022covid}. For classification, the larger the output sets $\Gamma_n^{1-\epsilon}$ are, the less precisely the model has predicted a sample $n$'s target $y_n$ \cite{nguyen2018cover}. Relevant CP evaluation metrics are defined in \Cref{tab:cpMetrics}. Note that metrics measured with the prediction sets are dependent on the significance level $\epsilon$.
        
        For CP approximation, as with our proposed loss function, validity is not guaranteed but instead approximated. As a result, characteristics related to the success of the approximation should be measured in addition to the standard CP metrics. This includes the Kolmogorov-Smirnov test for uniformity \cite{zhang2010fast}, the miscalibration rate (distance between the expected and achieved error curves), and Fuzziness~\cite{vovk2016criteria} to measure the p-value distributions.
        
        \begin{table}[!h]
            \centering
            \caption{Evaluation metrics for Conformal Prediction (CP) predictive efficiency, and distribution metrics to evaluate CP approximation.}\label{tab:cpMetrics}
            \begin{tabular}{@{\hskip6pt}>{}l@{\hskip6pt}>{}c@{\hskip6pt}>{}l}
                \toprule
                \bfseries Metric & \bfseries Formula & \bfseries Intuition \\
                \specialrule{.4pt}{2pt}{0pt}\midrule
                Error rate & $(\sum^{N}_{n=1}y_n \notin \Gamma_n^{1-\epsilon}) / N$ & True label not in prediction set. \\
                Empty rate & $(\sum_{n=1}^N \text{\textbar}  \Gamma_n^{1-\epsilon }\text{\textbar} = 0) / N$ & Outlier samples. \\
                Single rate & $(\sum^{N}_{n=1} \text{\textbar} \Gamma_n^{1-\epsilon }\text{\textbar} =1) / N$ & Efficient predictions. \\
                Multi rate & $(\sum^{N}_{n=1} \text{\textbar} \Gamma_n^{1-\epsilon }\text{\textbar} >1) / N$ & Samples on the class boundary. \\
                \midrule
                KS-test & $D = \max \lvert P - \mathcal{U}\rvert$ & Statistical uniformity test for distribution $P$. \\
                Miscalib. & $\lvert\sum error_{\epsilon} - \epsilon\rvert$ & Distance between error line and expected error. \\
                Fuzziness & $\frac{1}{k} \sum^{N}_{n=1} \left( \sum_y p_n^y - \max_y p_n^y \right) $ & Smaller values are preferable. \\
                \specialrule{.4pt}{2pt}{0pt}\bottomrule
            \end{tabular}
        \end{table}

    \subsection{Neural network architecture and hyperparameter optimisation}\label{sec:nnArchitecture}    
\begin{figure}[!h]
            \centering
            \includegraphics[width=0.4\textwidth]{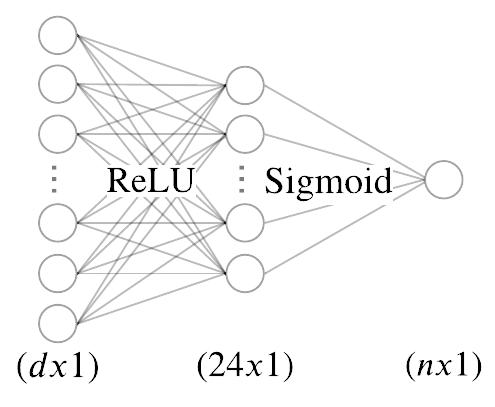}
            \caption{Neural network feedforward architecture. Our proposed conformal loss function requires the final activation to be sigmoid, and the number of output neurons to match the number of classes $n$. The number of input neurons $d$ varies based on the number of features, see \Cref{sec:dataset}.}\label{fig:nnArchitecture}
        \end{figure}

        {We opt for a shallow feedforward neural network (NN) architecture as shown in \Cref{fig:nnArchitecture}. Similar architectures have been previously successfully used for the classification of large datasets such as MNIST ~\cite{chazal2015feedforward,lejeune2020mechanical,gabella2020topology}, supporting our decision for a simpler classification model (e.g., compared to traditional Convolutional Neural Networks for images~\cite{kayed2020classification,kadam2020cnn,garg2019validation}). A test with the defined architecture, no hyperparameter optimisation, and standard Sparse Categorical Cross Entropy shows that the model has enough complexity to accurately represent and learn relationships of 7 classification tasks on all 5 datasets, see \Cref{tab:nnTestAccuracy}.}
        
        \begin{table}[!h]
            \centering\setlength{\tabcolsep}{2.3pt}
            \caption{Test accuracy of the neural network visualised in \Cref{fig:nnArchitecture} for the datasets described in \Cref{sec:dataset}. The model has enough complexity to accurately represent relationships between the input data and class labels.}\label{tab:nnTestAccuracy}
            \begin{tabular}{>{}r>{}r>{}r>{}r>{}r>{}r>{}r>{}r}
            \toprule
            & \textbf{MNIST2} & \textbf{MNIST10} & \textbf{USPS2} & \textbf{USPS10} & \textbf{BANK} & \textbf{WINE} & \textbf{MSHRM} \\
            \midrule
            \textbf{Accuracy} & 99.95\% & 95.12\% & 99.36\% & 89.54\% & 91.06\% & 93.80\% & 99.94\% \\
            \bottomrule
            \end{tabular}
        \end{table}
        
        There are two simple architecture requirements for compatibility with our conformal loss: The final activation must be a sigmoid function, which limits the output to the range (0, 1) (see \Cref{sec:cpBackground}); And the number of output neurons $n$ must be equal to the number of classes. Since we are interested in binary and multi-class classification, we use two models: one with $n=2$, and one with $n=10$.
        
\begin{figure}[!h]
            \centering
            \includegraphics[width=0.60\textwidth]{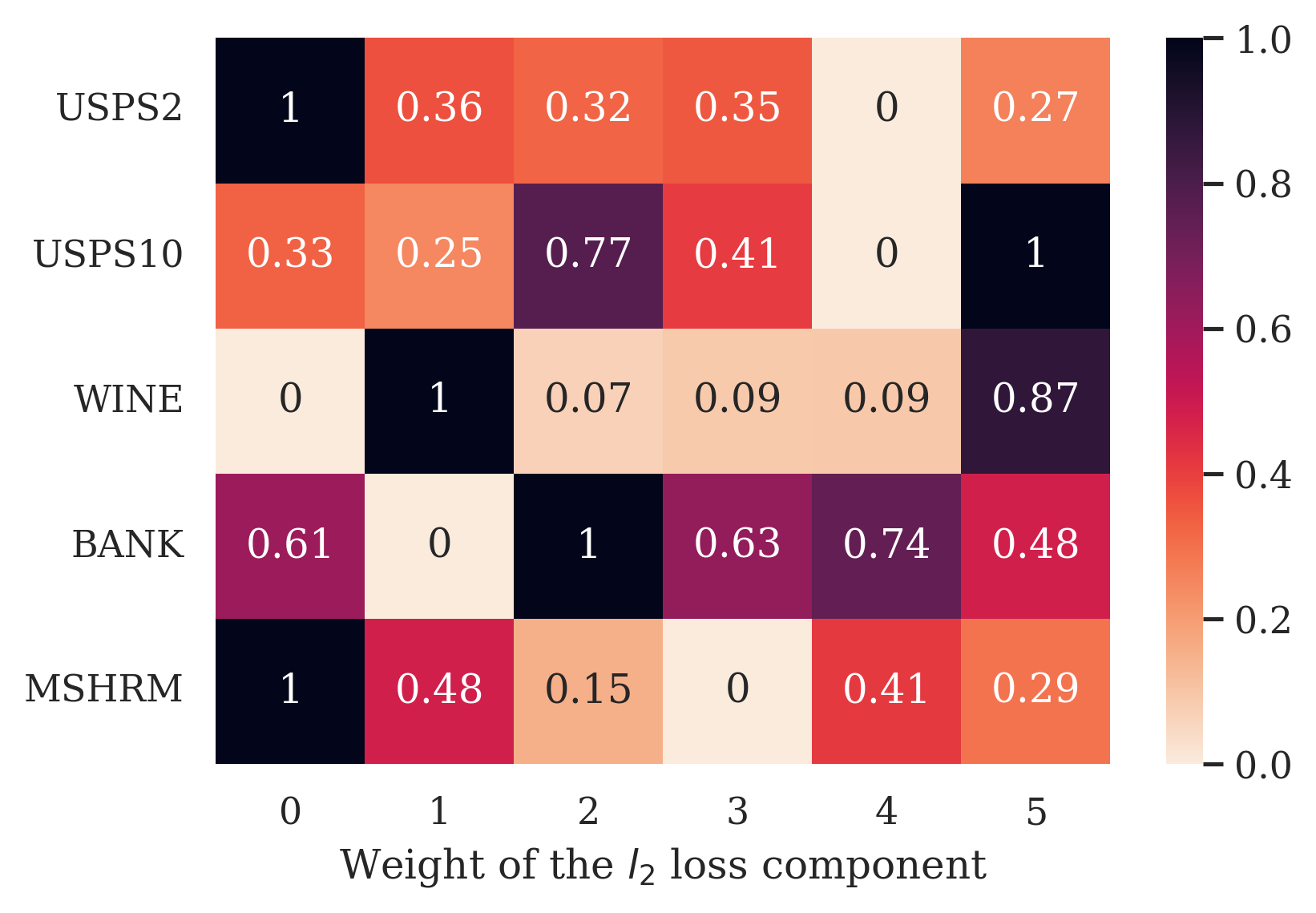}
            \caption{Normalised grid-search results for loss hyperparameter optimisation. The heat map shows the normalised miscalibration rate achieved by our CP approximation model on the four remaining datasets. The weight with the lowest value (0) is chosen for each dataset.}\label{fig:gridsearchResults}
        \end{figure}

        The model's optimal hyperparameters were identified with an extensive grid search based on the CP approximation output on the MNIST dataset (i.e., miscalibration, see \Cref{sec:evalMetrics}). The parameter values were as follows:
        \begin{itemize}
            \item For the NN: optimiser = adam, learning rate =  0.001, batch size = 128, epochs = 3.
            \item For the conformal loss function (see \Cref{eq:lossTotal,eq:lossTrue} in \Cref{sec:proposedMethod}):
            \begin{gather}
                loss = 1 \cdot loss_{false} + 1 \cdot loss_{true} \\
                \begin{aligned}
                    loss_{true} ={}  & 1 \cdot (loss_{mean} + loss_{var}) + 5 \cdot loss_{l2} \\
                      & +{} 0.25 \cdot loss_{huber}(\delta=0.125) \label{eq:lossParams}
                \end{aligned}
            \end{gather}
        \end{itemize}
        For the other 4 datasets, we carried out a narrower grid search. We found that the best-performing loss had the same weights for most components as reported in \Cref{eq:lossParams}. Only the $l_2$ component weight changed, as shown in \Cref{fig:gridsearchResults}.

    \subsection{Empirical evaluation of the proposed conformal loss function}
        For the most direct comparison between our conformal loss function (\Cref{sec:proposedMethod}), Aggregated Conformal Prediction (ACP), and Inductive Conformal Prediction (ICP), we use the same feedforward neural network (NN) model in all tests: as a standalone Deep Learning (DL) model and as the underlying model for ACP and ICP, since any point predictor may be used (\Cref{sec:CP}). The ACP p-value aggregation procedure and ICP non-conformity-measure are given in \Cref{eq:acpPvalue,eq:acpNCM} respectively. Readers are referred to \Cref{sec:cpBackground} for a detailed background on CP.
        
        Ten iterations were evaluated in all 3 scenarios. For DL and ICP, each iteration represents an independent run, and the average is reported; For ACP, each run had a different number of ensemble models from $2 \leq n \leq 10$. Each NN model was trained with our novel conformal loss function after parameter optimisation (\Cref{sec:results}). The models were evaluated with metrics suited to Conformal Prediction, as described in \Cref{sec:evalMetrics}. We start with a detailed overview of the binary classification results and then explore how the increase of classes and training samples affected the model performance. Overall, our analysis had a stronger focus on calibration adherence at lower significance levels $\epsilon \in \{0.05, 0.1, 0.2\}$, since low error rates are especially relevant to most prediction tasks with confidence requirements.

        To give a comprehensive results analysis, we first evaluate the MNIST dataset for the binary and multi-class tasks in great detail (\Cref{sec:mnistResults}), after which we highlight similarities and differences for the additional 4 datasets (\Cref{sec:otherResults}).

        \subsubsection{MNIST evaluation}\label{sec:mnistResults}
\begin{figure}[!h]
                \centering
                \begin{subfigure}{0.32\textwidth}
                    \centering
                    \includegraphics[width=\textwidth]{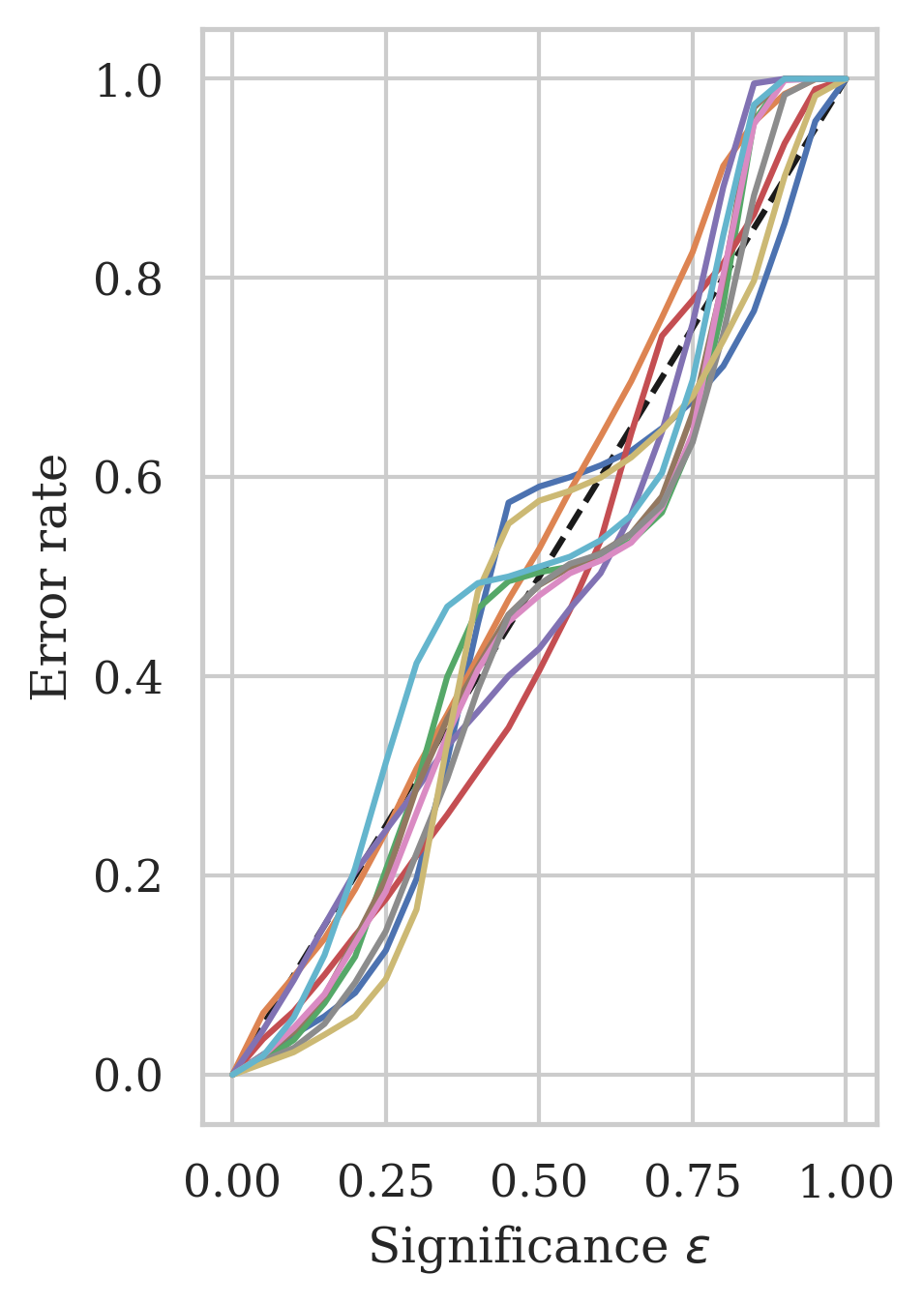}
                    \caption{10 NN iterations.}
                    \label{fig:NnCalibration}
                \end{subfigure}
                \hfill
                \begin{subfigure}{0.32\textwidth}
                    \centering
                    \includegraphics[width=\textwidth]{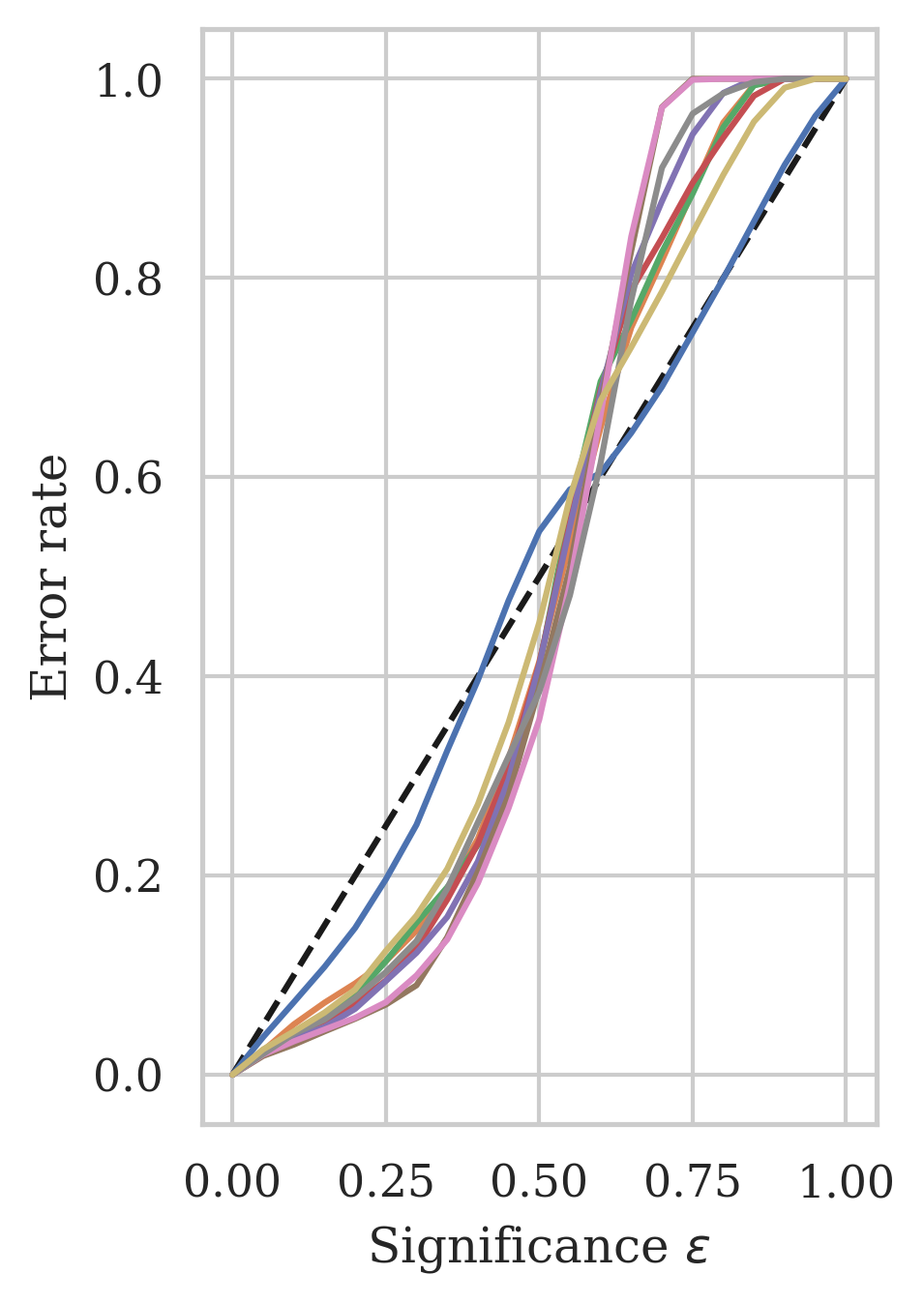}
                    \caption{ACP ($2 \leq n \leq 10$).}
                    \label{fig:AcpCalibration}
                \end{subfigure}
                \hfill
                \begin{subfigure}{0.32\textwidth}
                    \centering
                    \includegraphics[width=\textwidth]{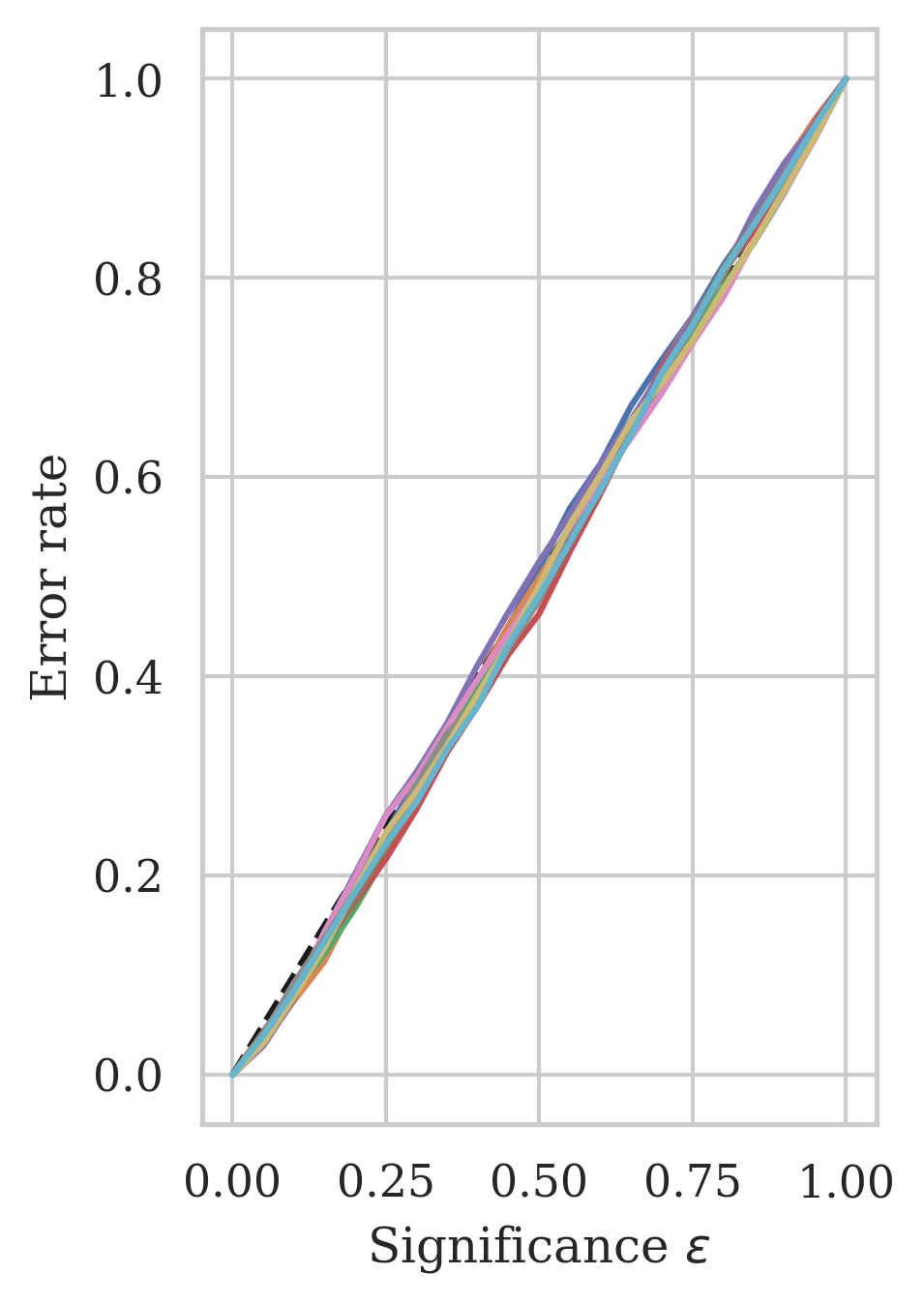}
                    \caption{10 ICP iterations.}
                    \label{fig:icpCalibration}
                \end{subfigure}
                \caption{The calibration curves to evaluate the approximate validity of NN models trained with our conformal loss function compared to ACP and ICP on MNIST2. Both methods approximation methods tend to be conservative for low significance levels.}\label{fig:modelCalibration}
            \end{figure}

            \Cref{fig:modelCalibration} confirms that the error rate of the CP approximation with our proposed NN loss function roughly followed the calibration curve (dashed line) for binary classification on the MNIST2 dataset. Unlike ACP, which with $n>2$ has a distinctive S-shaped curve (as previously noted in \cite{linusson2017calibration}), our proposal seemed to conform much more closely to the ideal validity curve at any given point. While NN error rates were conservative at low significance ($\epsilon<0.3$), the gap was overall smaller than ACP with $n>2$.
            
            Hence, the results confirm that our conformal loss achieved approximate validity without computing intermediate non-conformity measures. The calibration curve deviations and their concrete effects will be examined in more detail when we evaluate the calibration/efficiency trade-off later in this section.
            
\begin{figure}[!h]
                \centering
                \begin{subfigure}{0.45\textwidth}
                    \centering
                    \includegraphics[width=\textwidth]{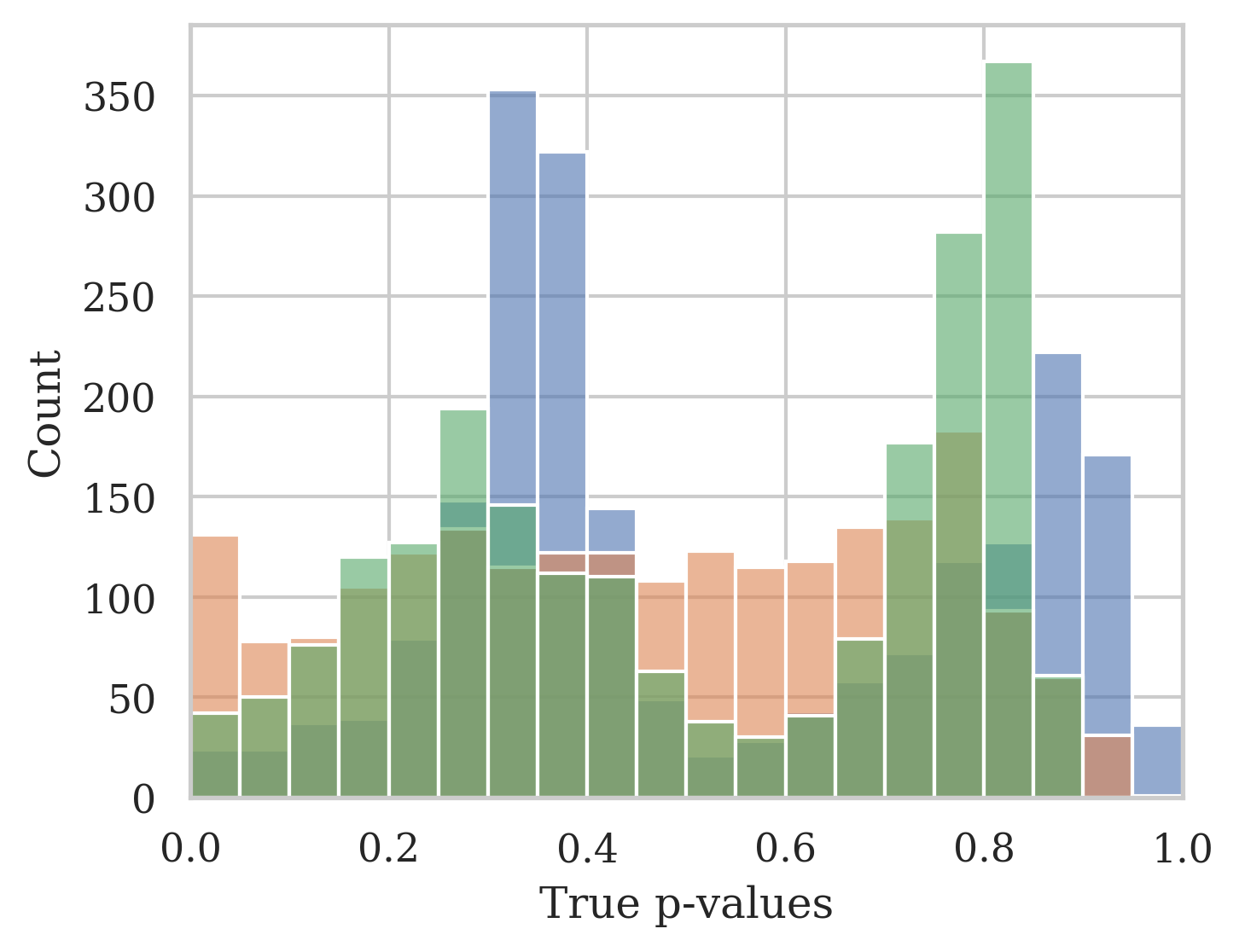}
                    \caption{NN.}
                    \label{fig:NnTruePvalues}
                \end{subfigure}
                \hfill
                \begin{subfigure}{0.45\textwidth}
                    \centering
                    \includegraphics[width=\textwidth]{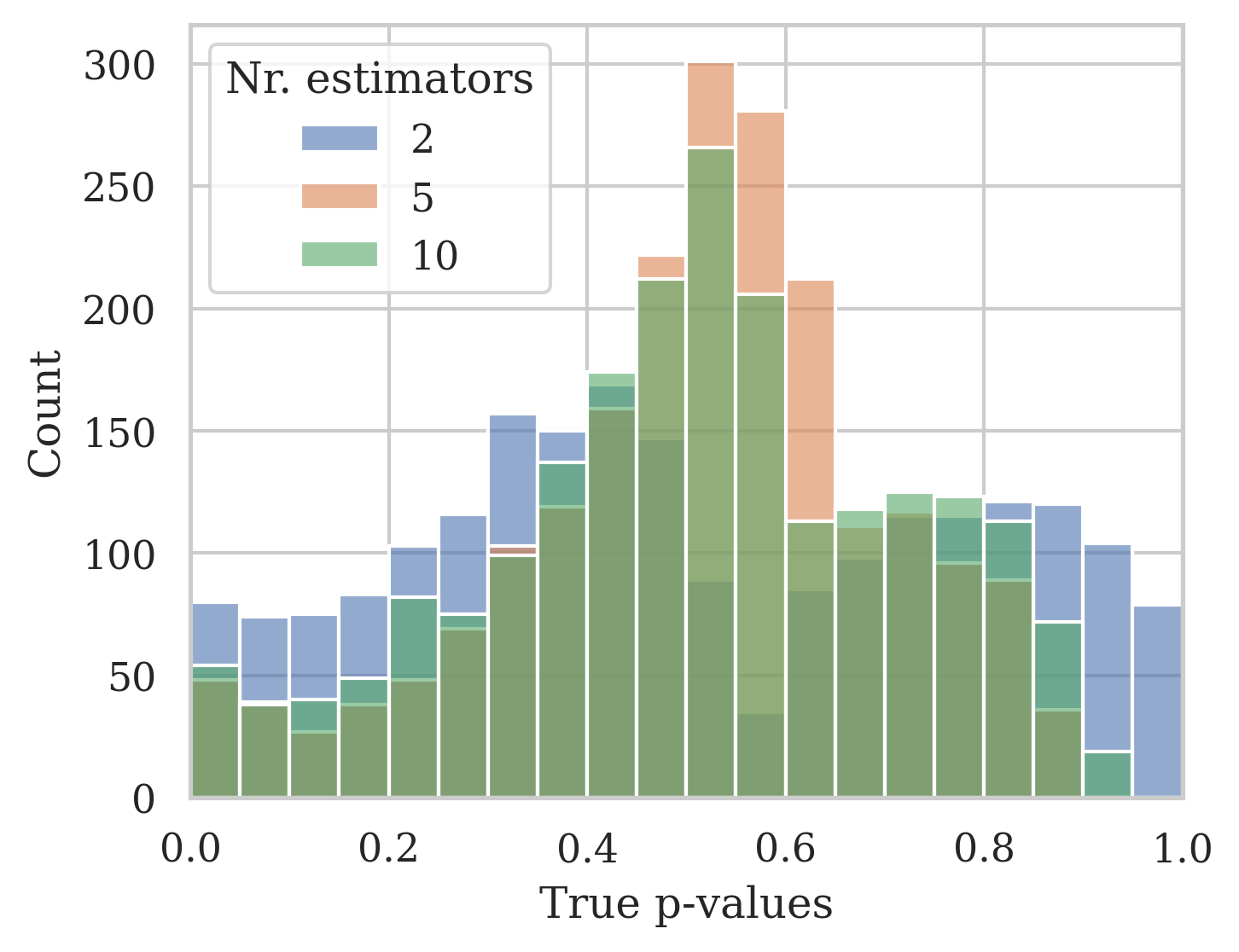}
                    \caption{ACP.}
                    \label{fig:AcpTruePvalues}
                \end{subfigure}
                \hfill
                \begin{subfigure}{0.45\textwidth}
                    \centering
                    \includegraphics[width=\textwidth]{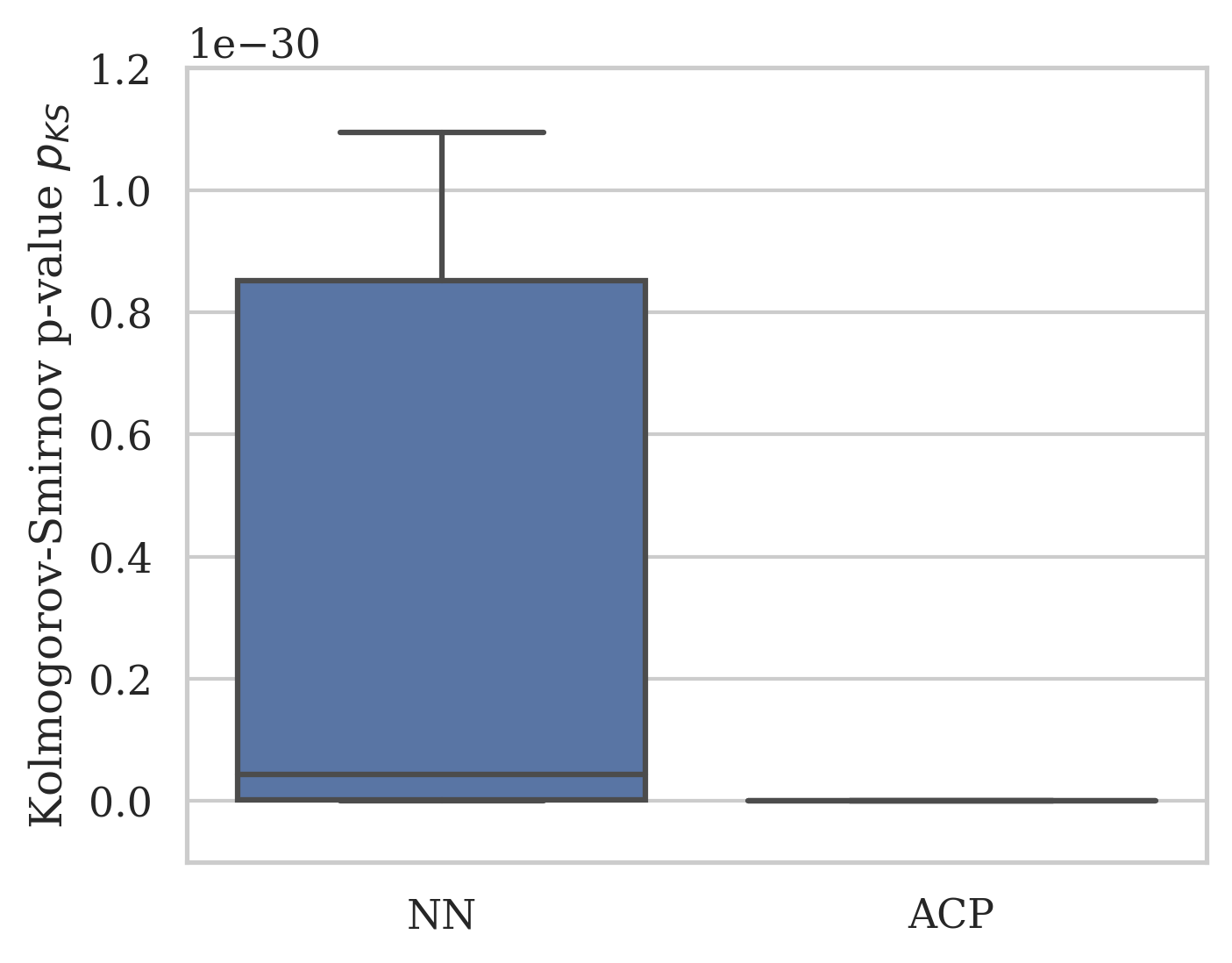}
                    \caption{Uniformity test.}
                    \label{fig:uniformityTest}
                \end{subfigure}
                \caption{True p-value distribution of three randomly selected NN models and three representative CP models on MNIST2. Neither model type achieves the expected uniform distribution $\mathcal{U}_{[0,1]}$, which is statistically confirmed with the Kolmogorov-Smirnov test for uniformity.}\label{fig:truePvalues}
            \end{figure}

            \paragraph{Approximate validity}
            To preserve validity, p-values of the true class should follow the uniform distribution $\mathcal{U}_{[0, 1]}$. On the other hand, false-class p-values should be as close to 0 as possible to maximise predictive efficiency (\Cref{sec:proposalRequirements}). An NN trained with our conformal loss function tended to a bi-modal distribution of true-class p-values, dipping around 0.5 (\Cref{fig:truePvalues}), which may be a side-effect of the Huber loss component. Inversely, ACP tended to a unimodal distribution, peaking at 0.5 (\Cref{fig:AcpTruePvalues}). However, we note that $n=2$ was also bimodal and distinctly similar to the NN trends.
            
            These findings are directly related to the calibration lines and explain previous observations in \Cref{fig:calibration}: NN with true p-values concentrated towards the extremes tended to be slightly conservative at lower ($\epsilon < 0.3$) and higher ($\epsilon > 0.7$) significance levels. In contrast, ACP $n>1$ had a distinctive S-shape and was significantly conservative until $\epsilon=0.5$, after which the calibration curve crossed the diagonal and became invalid.
            
            Neither method passed the Kolmogorov-Smirnov test of uniformity \cite{zhang2010fast}, since all p-values $p_{KS}$ were lower than our significance level $\alpha_{KS}=0.01$ (\Cref{fig:uniformityTest}). However, NN $p_{KS}$ were distributed across a larger range than ACP $n \geq 2$ ($p_{KS}$ has an inverse relationship with $n$), which shows that our method has the potential to achieve stronger approximate validity with future loss function optimisations.
            
\begin{figure}[!h]
                 \centering
                 \begin{subfigure}{0.32\textwidth}
                     \centering
                     \includegraphics[width=\textwidth]{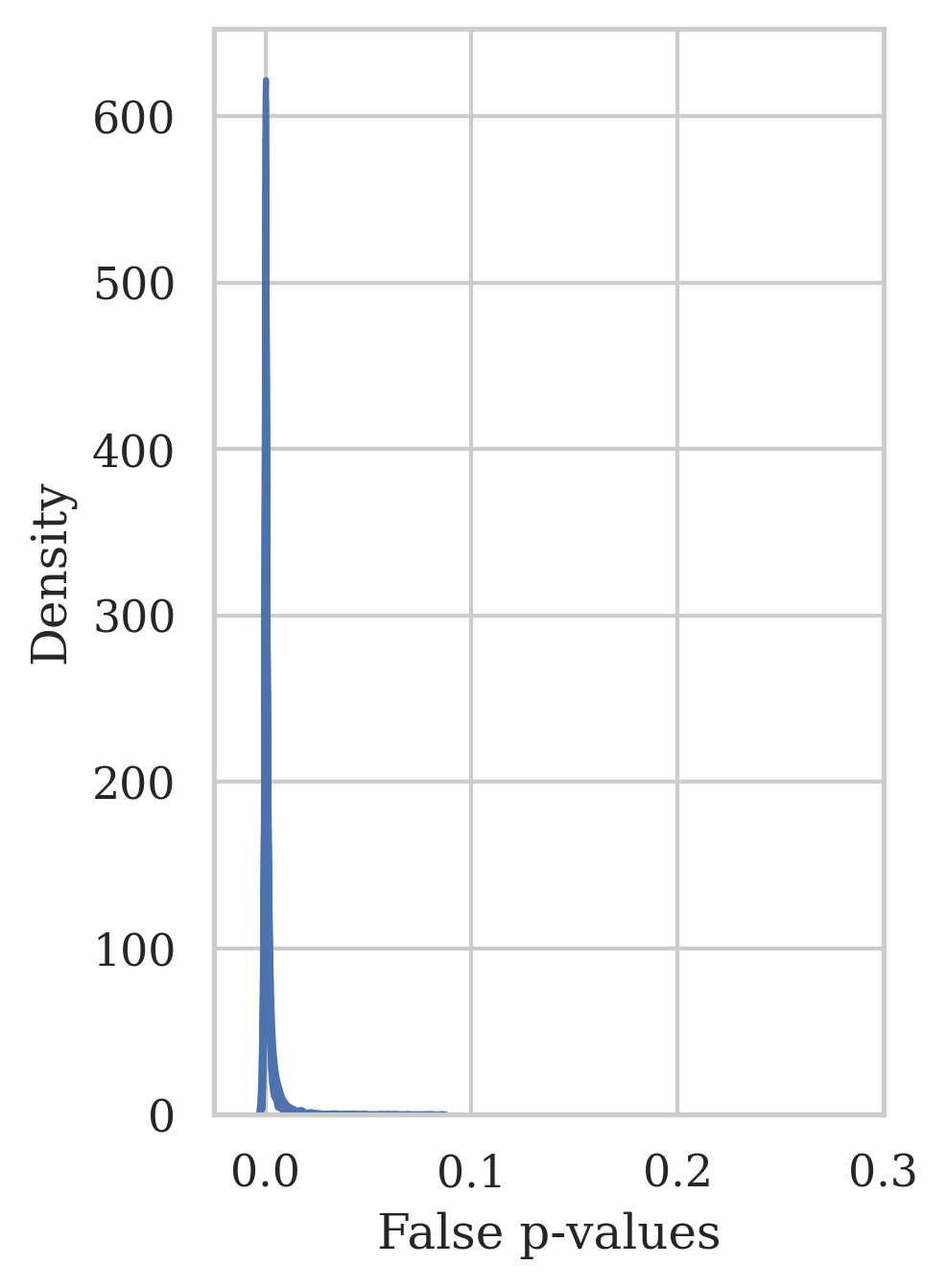}
                     \caption{10 NN iterations.}
                     \label{fig:NnFalsePvalues}
                 \end{subfigure}
                 \hfill
                 \begin{subfigure}{0.32\textwidth}
                     \centering
                     \includegraphics[width=\textwidth]{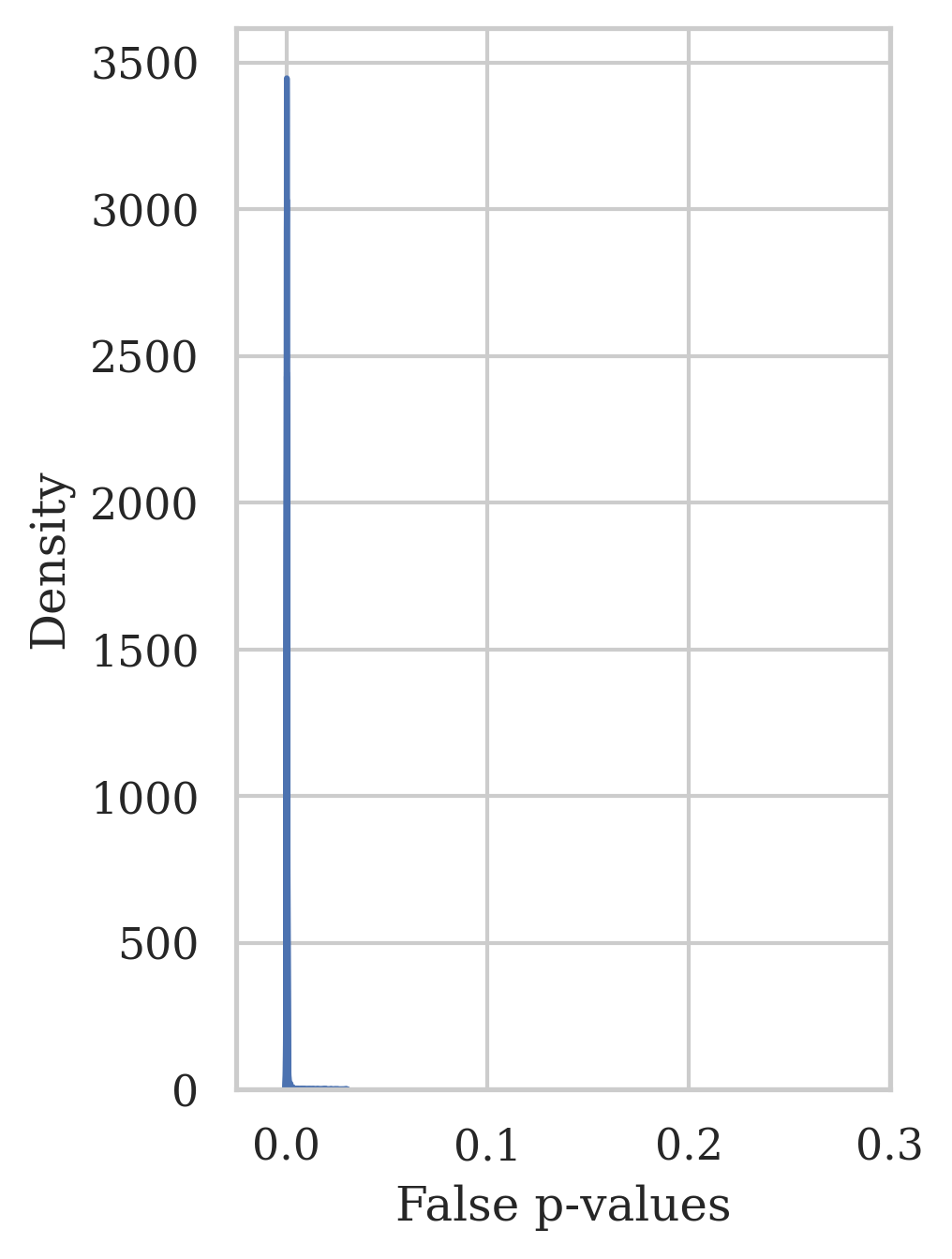}
                     \caption{ACP ($2 \leq n \leq 10$).}
                     \label{fig:AcpFalsePvalues}
                 \end{subfigure}
                 \hfill
                 \begin{subfigure}{0.32\textwidth}
                     \centering
                     \includegraphics[width=\textwidth]{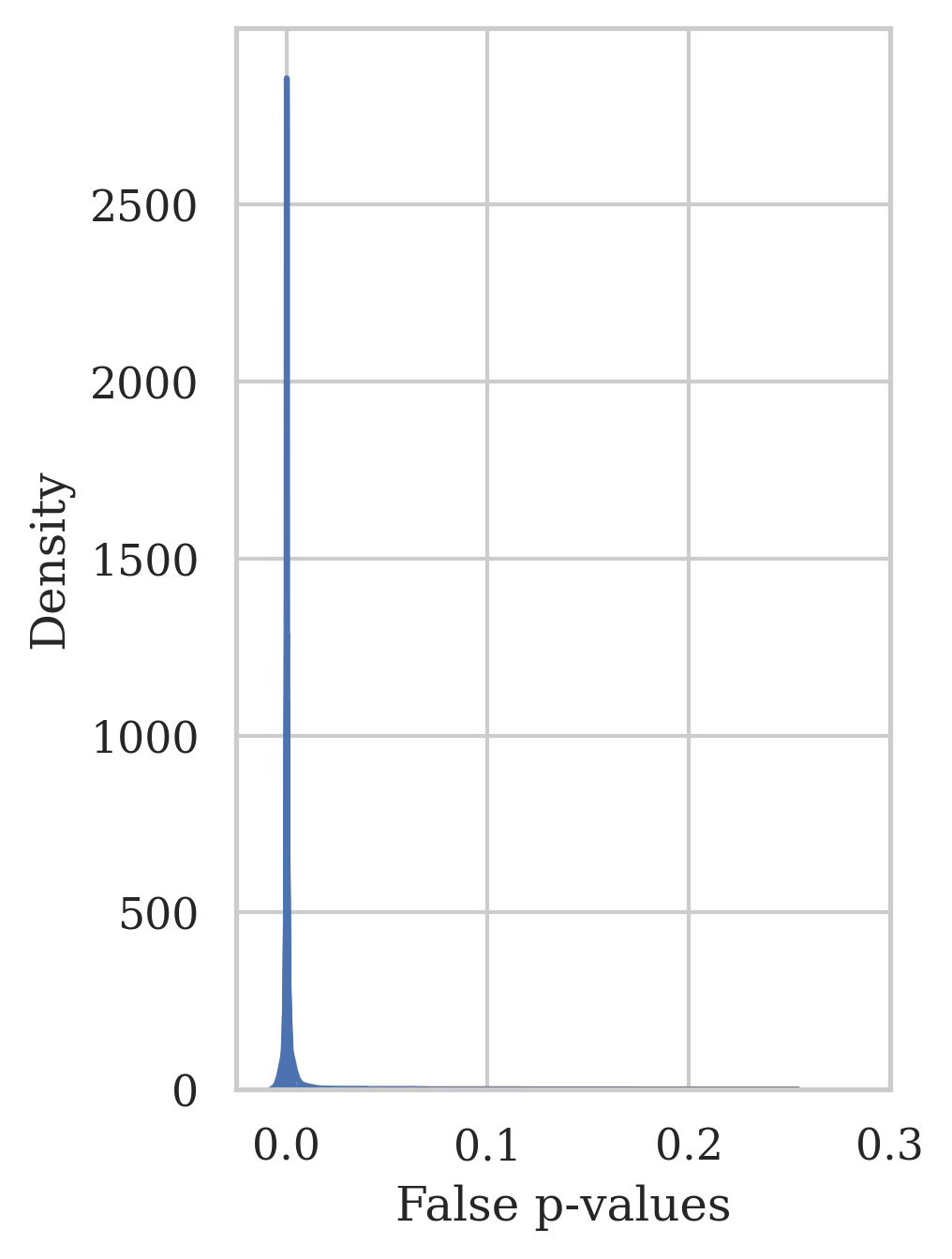}
                     \caption{10 ICP iterations.}
                     \label{fig:IcpFalsePvalues}
                 \end{subfigure}
                 \caption{False p-value distribution of the NN, ACP, and ICP model types. P-values produced by our conformal loss function are negligibly larger than ACP and ICP. However, the predictive efficiency is not affected, since they remain smaller than the lowest considered significance level $\epsilon=0.05$.}\label{fig:falsePvalues}
            \end{figure}

            \paragraph{Predictive efficiency}
            \Cref{fig:falsePvalues} shows that the false-class p-values were centred around 0 for both methods. Surprisingly, increasing the number of models $n$ for ACP did not seem to improve predictive efficiency in this case, as all distribution curves were almost indistinguishably overlapped (\Cref{fig:AcpFalsePvalues}). While NN false p-values had slightly more spread up to 0.0005 (\Cref{fig:NnFalsePvalues}), the values were still orders of magnitude under the minimum $\epsilon = 0.05$ threshold considered in this study, and therefore did not affect the prediction set sizes. 
        
            \paragraph{Calibration vs predictive efficiency trade-off}
            After examining the p-value distributions, we may more meaningfully compare our proposed one-step conformal p-value approximation to traditional ACP on a more holistic level. We evaluated the predictive efficiency gains in the context of the distance of the error line to the calibration curve.
            
\begin{figure}[!h]
                \centering
                \begin{subfigure}{0.49\textwidth}
                    \centering
                    \includegraphics[width=\textwidth]{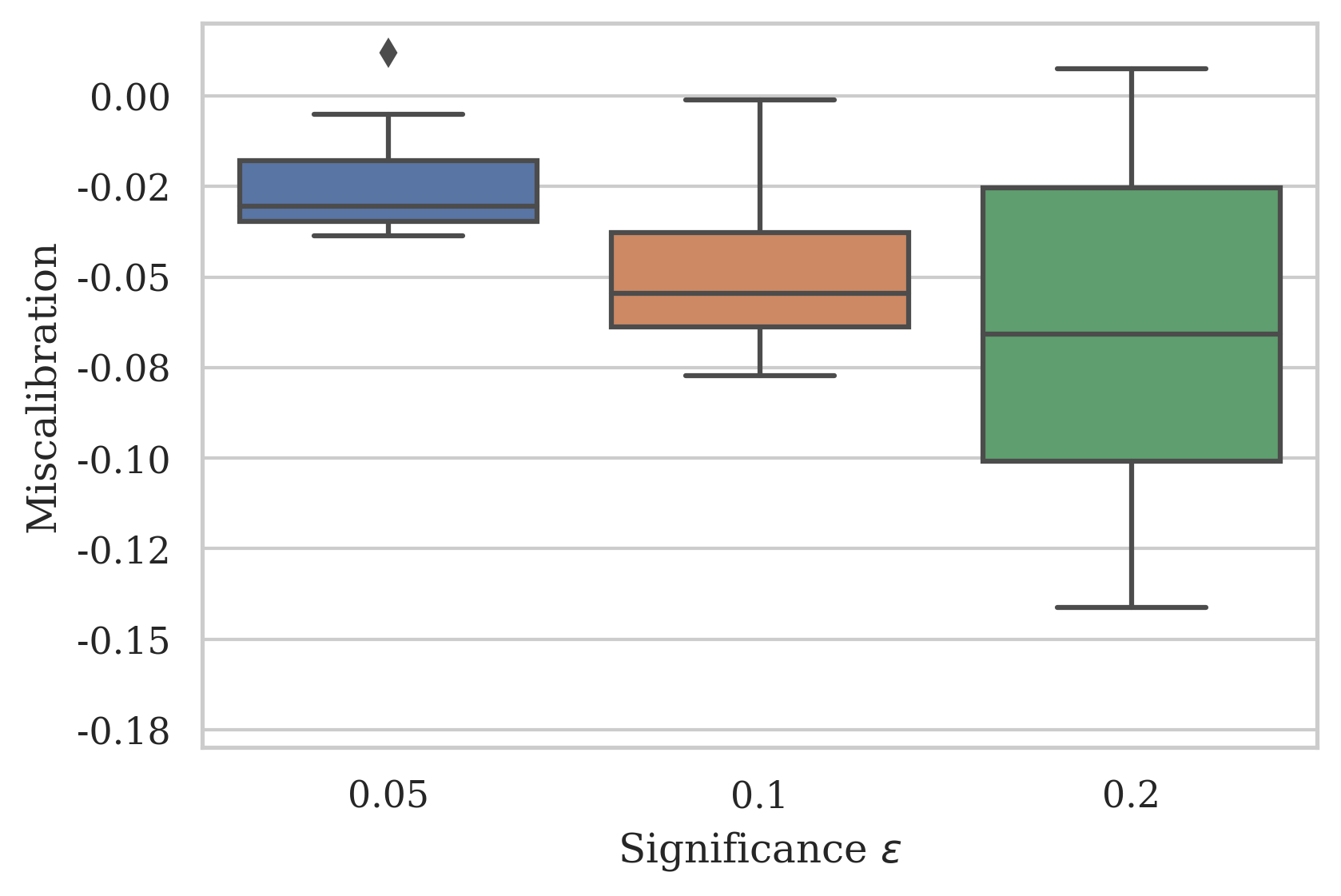}
                    \caption{NN miscalibration rate.}
                    \label{fig:NnCalibrationTrend}
                \end{subfigure}
                \hfill
                \begin{subfigure}{0.49\textwidth}
                    \centering
                    \includegraphics[width=\textwidth]{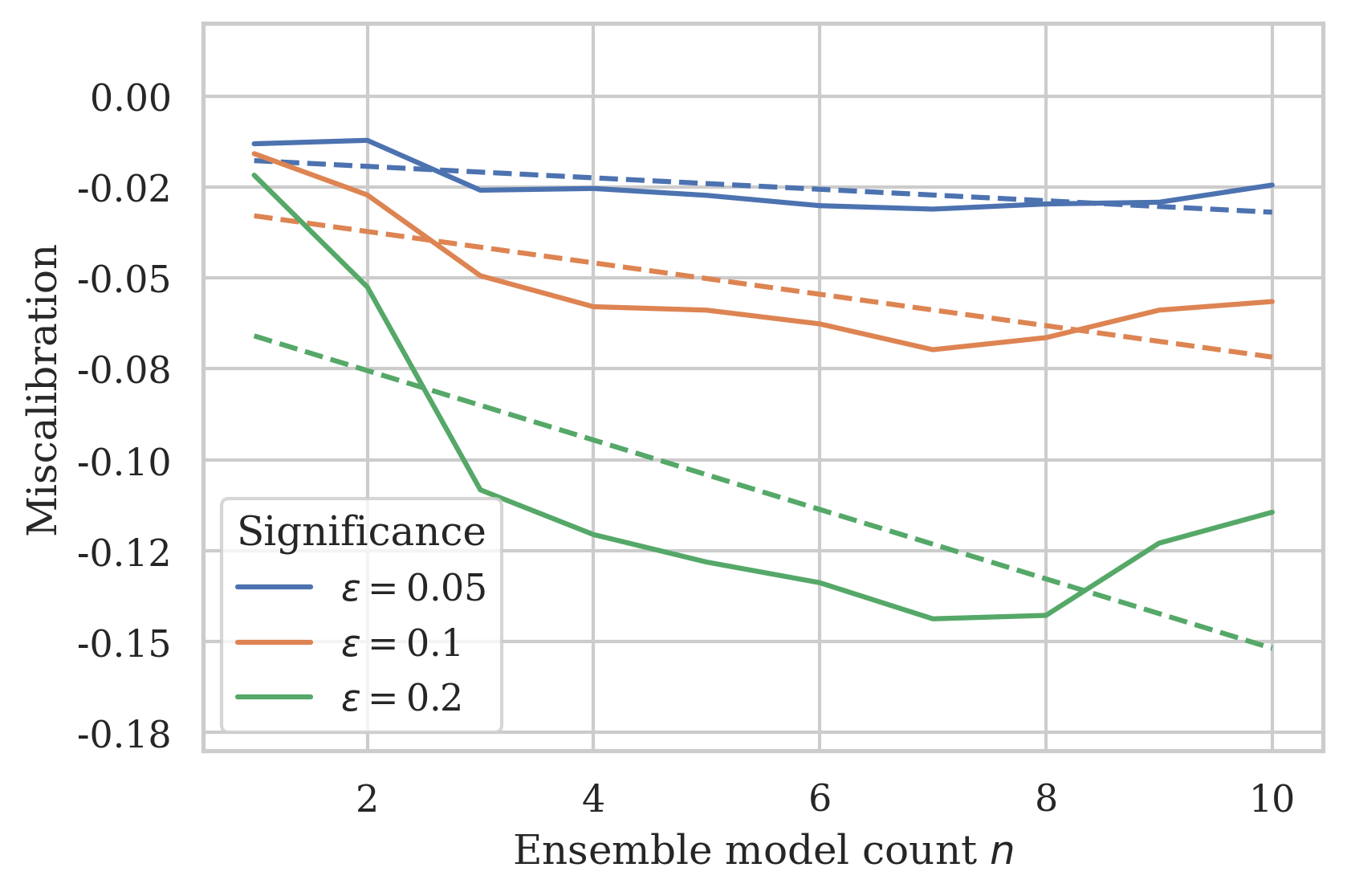}
                    \caption{ICP and ACP miscalibration rate.}
                    \label{fig:AcpCalibrationTrend}
                    
                \end{subfigure}
                \begin{subfigure}{0.49\textwidth}
                    \centering
                    \includegraphics[width=\textwidth]{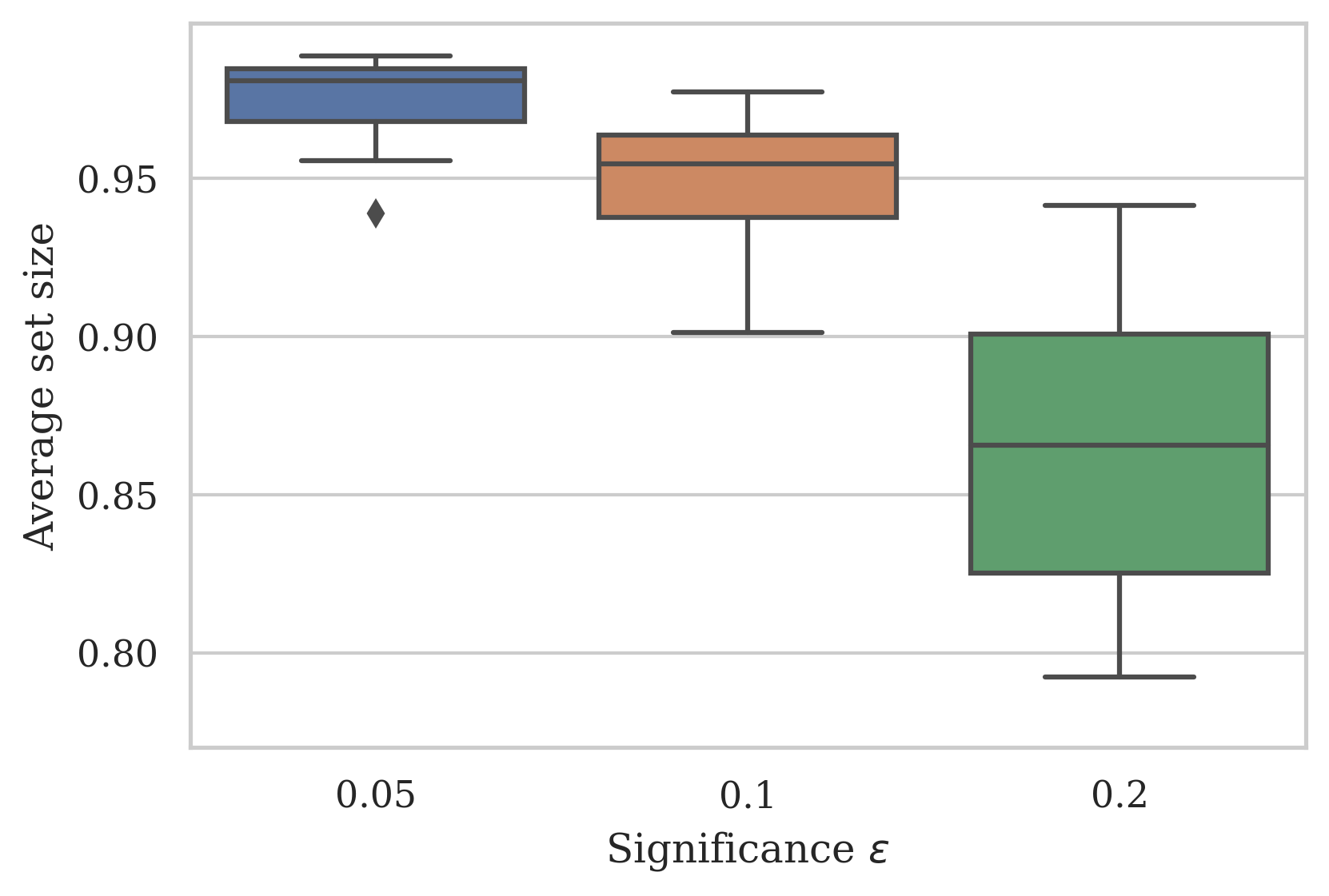}
                    \caption{NN average prediction set size.}
                    \label{fig:NnEfficiency}
                \end{subfigure}
                \hfill
                \begin{subfigure}{0.49\textwidth}
                    \centering
                    \includegraphics[width=\textwidth]{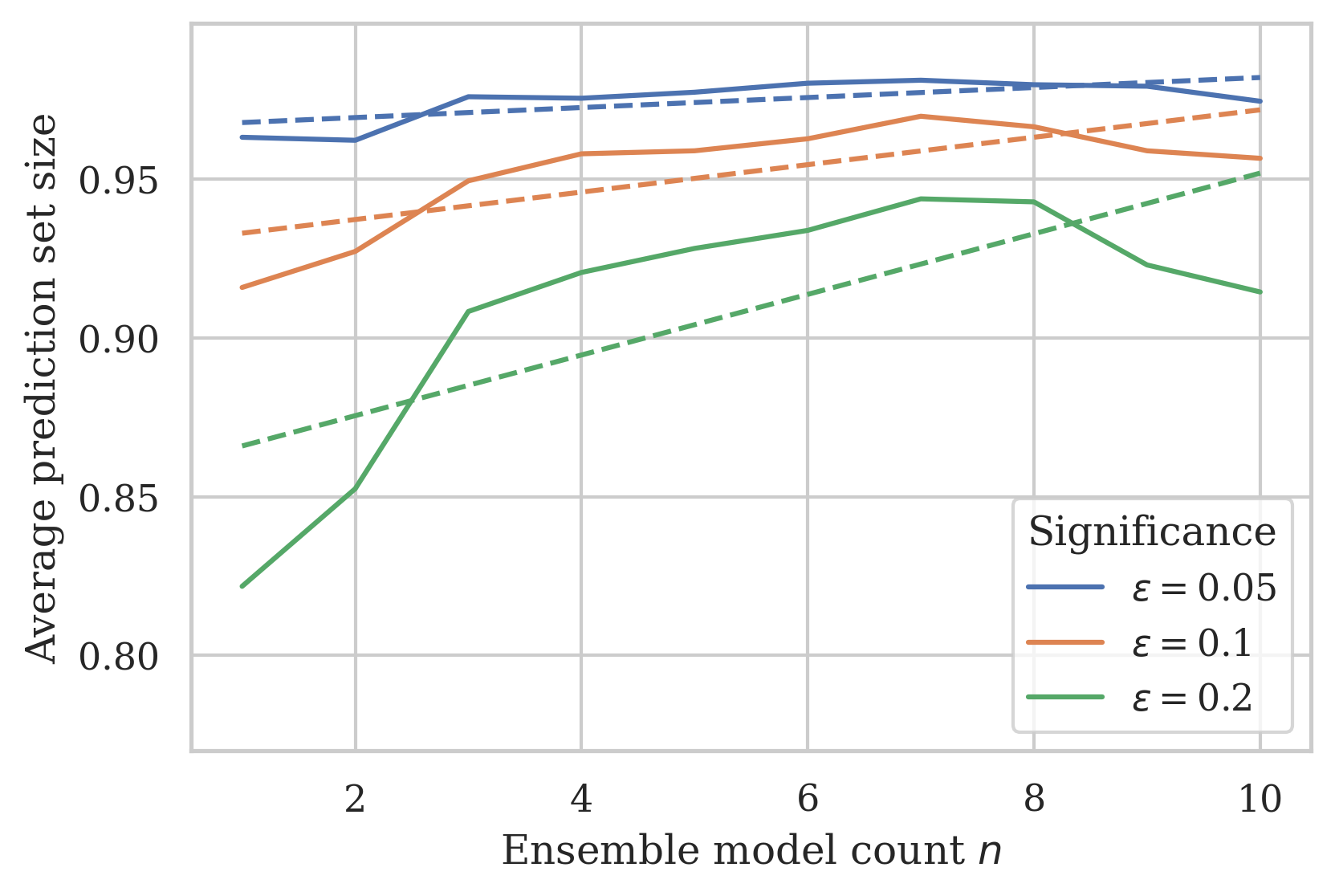}
                    \caption{ICP and ACP avg. prediction set size.}
                    \label{fig:AcpEfficiency}
                \end{subfigure}
                \caption{Calibration vs predictive efficiency trade-off for MNIST2, graphs are horizontally aligned with shared y-axes. Note that the optimal miscalibration rate is 0, and the optimal predictive efficiency is achieved with sets including exactly one label. Note that in (b) and (c), the first entry for $n=1$ corresponds to ICP. Interestingly, the competitiveness of our method increased with the significance level. As $\epsilon$ grew, the average calibration improvement of our model increased and was on par with ACP at lower $n$.}\label{fig:calibEffTradeoff}
            \end{figure}

            As discussed in \Cref{sec:CP}, ACP improves predictive efficiency by increasing set sizes towards the optimal as the number of ensemble models grows, with the side-effect of a weaker validity approximation (\Cref{fig:AcpCalibrationTrend}). Although the effects were relatively minor for small significance levels ($\epsilon= 0.05$), the positive efficiency and negative calibration trends became much more pronounced as $\epsilon$ grew.

            Promisingly, the average calibration of our conformal loss NN with $\epsilon = 0.05$ was competitive with the equivalent ACP trend line across all $n$, and the range improved the distance towards 0 in some iterations (\Cref{fig:NnCalibrationTrend}). At $\epsilon = 0.1$, NN again improved on ACP, with the mean equivalent to $n=5$ and some iterations reaching nearly perfect calibration (calibration distance $\approx$ 0). The largest improvement on average was achieved at $\epsilon = 0.2$, which on average significantly outperformed the ACP equivalent for $n>2$.
            
            In predictive efficiency, we may confirm again that our proposed loss function is competitive with ACP without major deficits. CP's optimal prediction set size is 1, and \Cref{fig:NnEfficiency,fig:AcpEfficiency} highlight that both models performed well, especially at low significance levels (NN=0.97 and ACP=0.96--0.97 for $\epsilon=0.05$. As expected, ACP average set sizes improved as the number of ensemble models $n$ increased. Similarly to the calibration distance, the trend slopes were shallower at low significance levels, and taking the calibration trade-off by increasing $n$ had a higher value of return at higher significance levels. Similarly, the NN model's predictive efficiency also showed larger ranges and variability as significance levels increased. This supports our previous assessment that the marginally increased p-values compared to ACP do not affect predictive efficiency.
            
            Standard CP performance metrics in \Cref{tab:setResults} confirm our observations so far. Both NN and ACP were conservative for low significance levels, although ACP achieved error scores closer to the expected maximum. Additionally, the minor increase in NN false p-values was negligible even at $\epsilon=0.05$, all significance levels show 0\% multi-sets and a high single-set rate. In combination with lower error rates, we conclude that our proposed conformal loss function successfully trains a simple NN model to confidently assign only the correct class to the vast majority of samples, without first calculating an additional non-conformity measure (\Cref{sec:proposedMethod}).
            
            \begin{table}[!h]
                \centering\setlength{\tabcolsep}{2pt}
                \caption{Overall Conformal Prediction approximation performances. NN and ICP results are averaged across 10 iterations. $n$ is the number of ACP ensemble classifiers. Results are given in \%.}\label{tab:setResults}
                \begin{tabular}{>{}l@{\hskip9pt}>{}r>{}r>{}r@{\hskip9pt}>{}r>{}r>{}r@{\hskip9pt}>{}r>{}r>{}r@{\hskip9pt}>{}r>{}r>{}r}
                    \toprule
                    & \multicolumn{3}{c}{\textbf{NN}} & \multicolumn{3}{c}{\textbf{ACP, $n=5$}} & \multicolumn{3}{c}{\textbf{ACP, $n=10$}} & \multicolumn{3}{c}{\textbf{ICP}} \\
                    \cmidrule(rr){2-4} \cmidrule(rr){5-7} \cmidrule(rr){8-10} \cmidrule(rr){11-13}
                    $\epsilon$ & \multicolumn{1}{c}{\textbf{0.05}} & \multicolumn{1}{c}{\textbf{0.1}} & \multicolumn{1}{c}{\textbf{0.2}} & \multicolumn{1}{c}{\textbf{0.05}} & \multicolumn{1}{c}{\textbf{0.1}} & \multicolumn{1}{c}{\textbf{0.2}} & \multicolumn{1}{c}{\textbf{0.05}} & \multicolumn{1}{c}{\textbf{0.1}} & \multicolumn{1}{c}{\textbf{0.2}} & \multicolumn{1}{c}{\textbf{0.05}} & \multicolumn{1}{c}{\textbf{0.1}} & \multicolumn{1}{c}{\textbf{0.2}}\\
                    \specialrule{.4pt}{2pt}{0pt}\midrule
                    \textbf{Error} & 2.61 & 5.31 & 13.58 & 4.49 & 9.50 & 20.38 & 1.89 & 5.77 & 20.76 & 3.56 & 8.28 & 18.37 \\
                    \textbf{Empty} & 2.59 & 5.31 & 13.58 & 4.44 & 9.50 & 20.38 & 1.89 & 5.77 & 20.76 & 3.42 & 8.23 & 18.36 \\
                    \textbf{Single} & 97.40 & 94.69 & 86.42 & 95.56 & 90.50 & 79.62 & 98.11 & 94.23 & 79.24 & 96.58 & 91.77 & 81.64 \\
                    \textbf{Multi} & 0.00 & 0.00 & 0.00 & 0.00 & 0.00 & 0.00 & 0.00 & 0.00 & 0.00 & 0.00 & 0.00 & 0.00 \\
                    \bottomrule
                \end{tabular}
            \end{table}

\begin{figure}[!h]
                \centering
                \begin{subfigure}{0.49\textwidth}
                    \centering
                    \includegraphics[width=\textwidth]{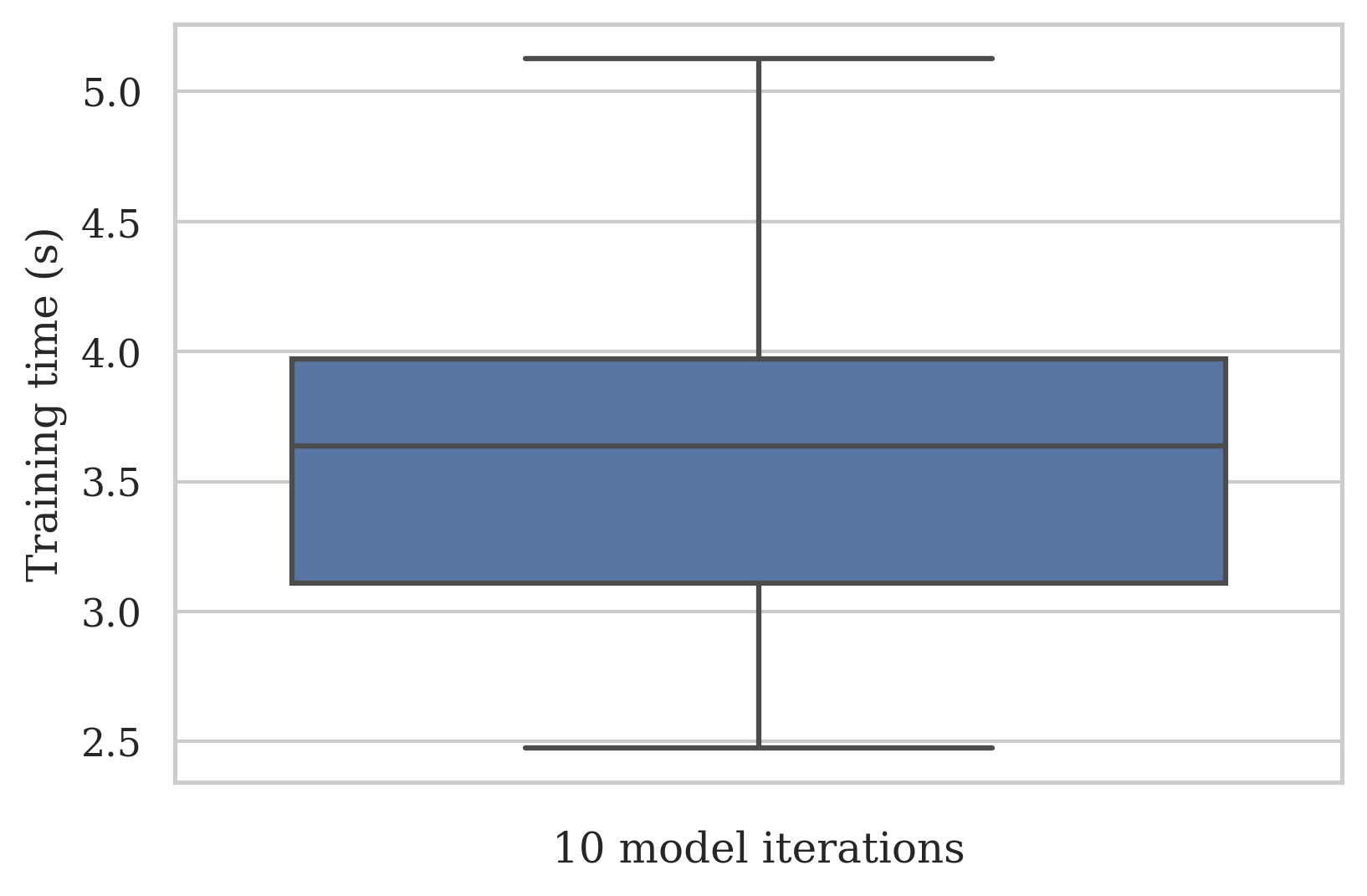}
                    \caption{NN.}
                    \label{fig:NnTrainingTime}
                \end{subfigure}
                \hfill
                \begin{subfigure}{0.49\textwidth}
                    \centering
                    \includegraphics[width=\textwidth]{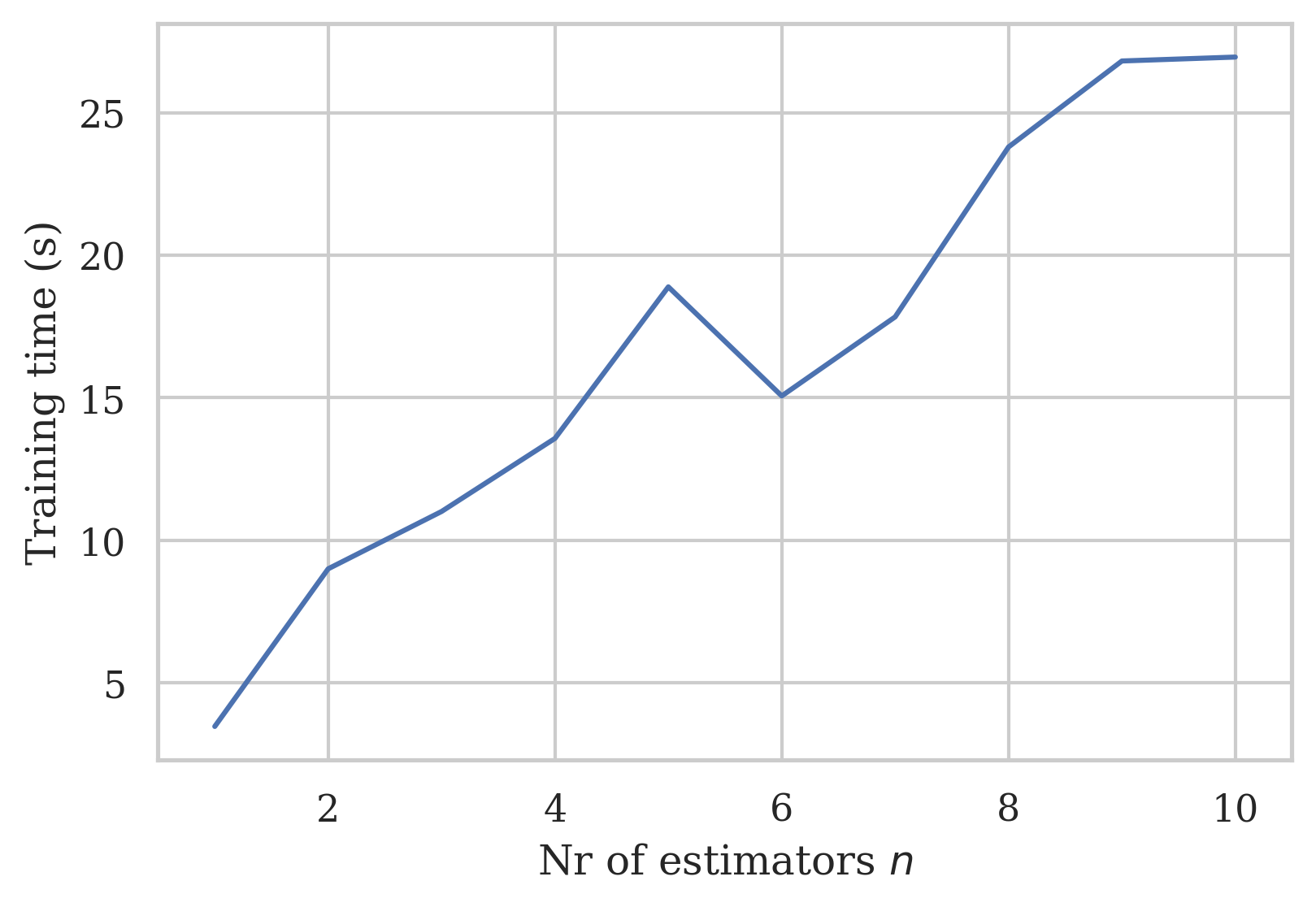}
                    \caption{ICP and ACP.}
                    \label{fig:AcpTrainingTime}
                \end{subfigure}
                \caption{Training times for the NN, ICP ($n=1$ in (b)), and ACP models. NN has a significant benefit over ACP, as only one model is trained in each iteration. The training time reduction by NN grows as the number of ensemble models increases, up to 86\% when $n=10$.}\label{fig:trainingTimes}
            \end{figure}

            \pagebreak
            \paragraph{Computational efficiency}
            After confirming that our proposed conformal loss function successfully maintains approximate validity and predictive efficiency on par with ACP for $3 \leq n \leq 10$, we evaluate the two methods' computational efficiency in \Cref{fig:trainingTimes}. This is where our model has two advantages: It skips the intermediate non-conformity measure calculation and requires training only one model. In contrast, ACP's training time increases linearly as the number of ensemble models $n$ increases. As a consequence, all test iterations of our proposed model significantly outperformed ACP with $n>1$ (3.5 seconds on average compared to up to over 25 seconds). Depending on $n$ and parallel computing capabilities, the training time gap may be narrowed, but ACP would nonetheless require significantly more computational power to train its ensemble models.
            
\begin{figure}[!h]
                \centering
                \begin{subfigure}{0.49\textwidth}
                    \centering
                    \includegraphics[width=\textwidth]{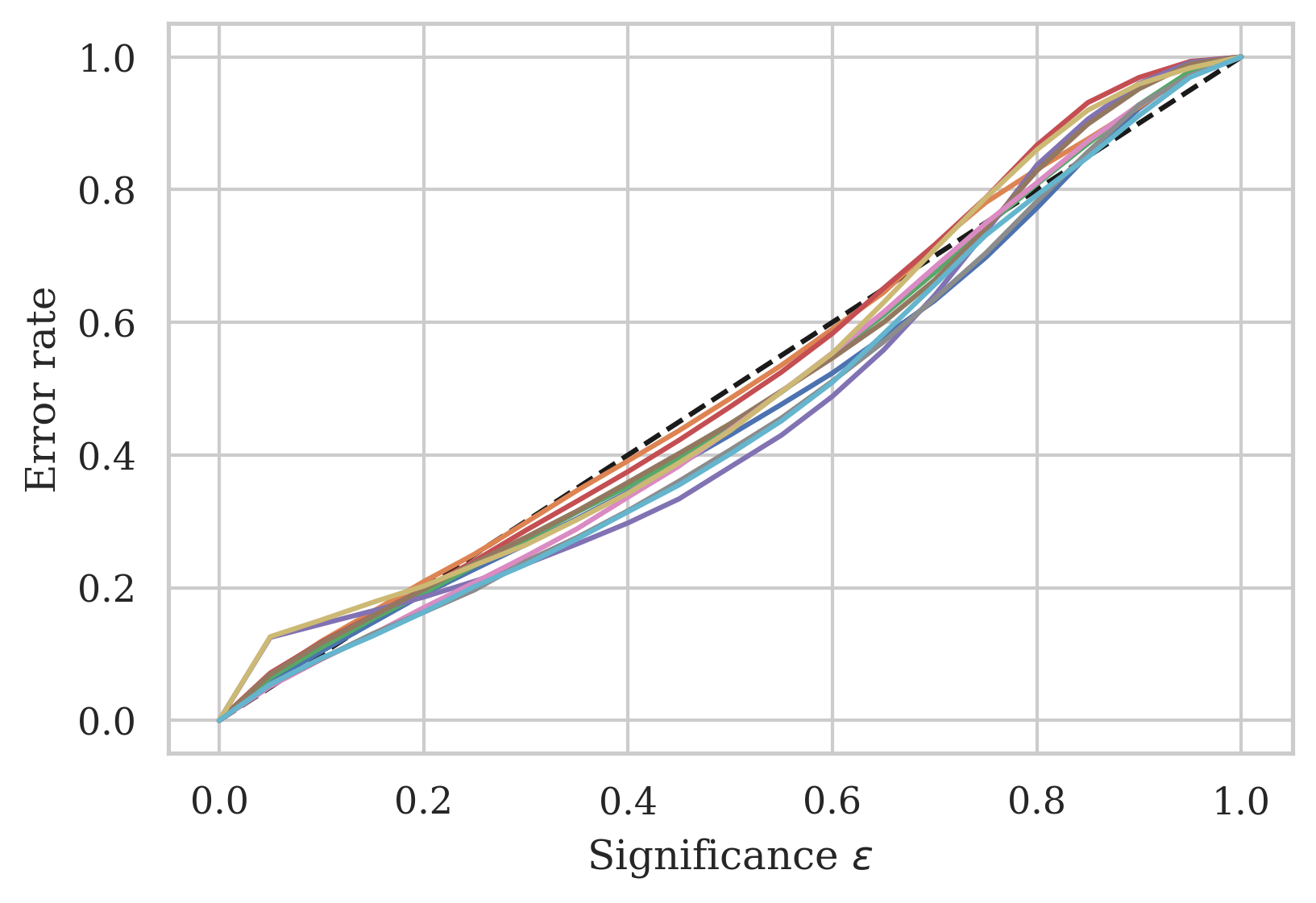}
                    \caption{NN calibration.}
                    \label{fig:nnCalibCurve10}
                \end{subfigure}
                \hfill
                \begin{subfigure}{0.49\textwidth}
                    \centering
                    \includegraphics[width=\textwidth]{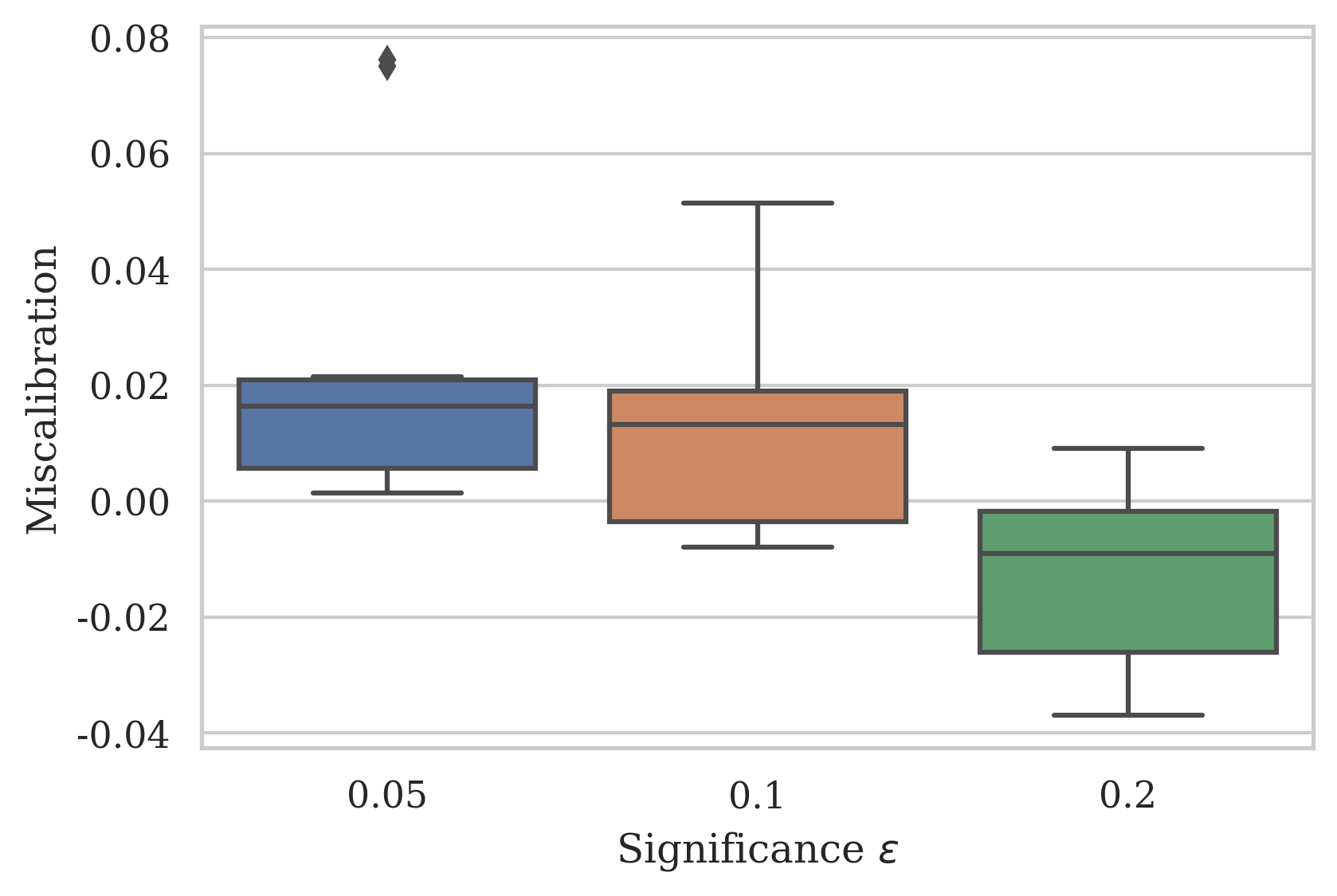}
                    \caption{NN calibration distance.}
                    \label{fig:nnDistTrend10}
                \end{subfigure}
                \begin{subfigure}{0.49\textwidth}
                    \centering
                    \includegraphics[width=\textwidth]{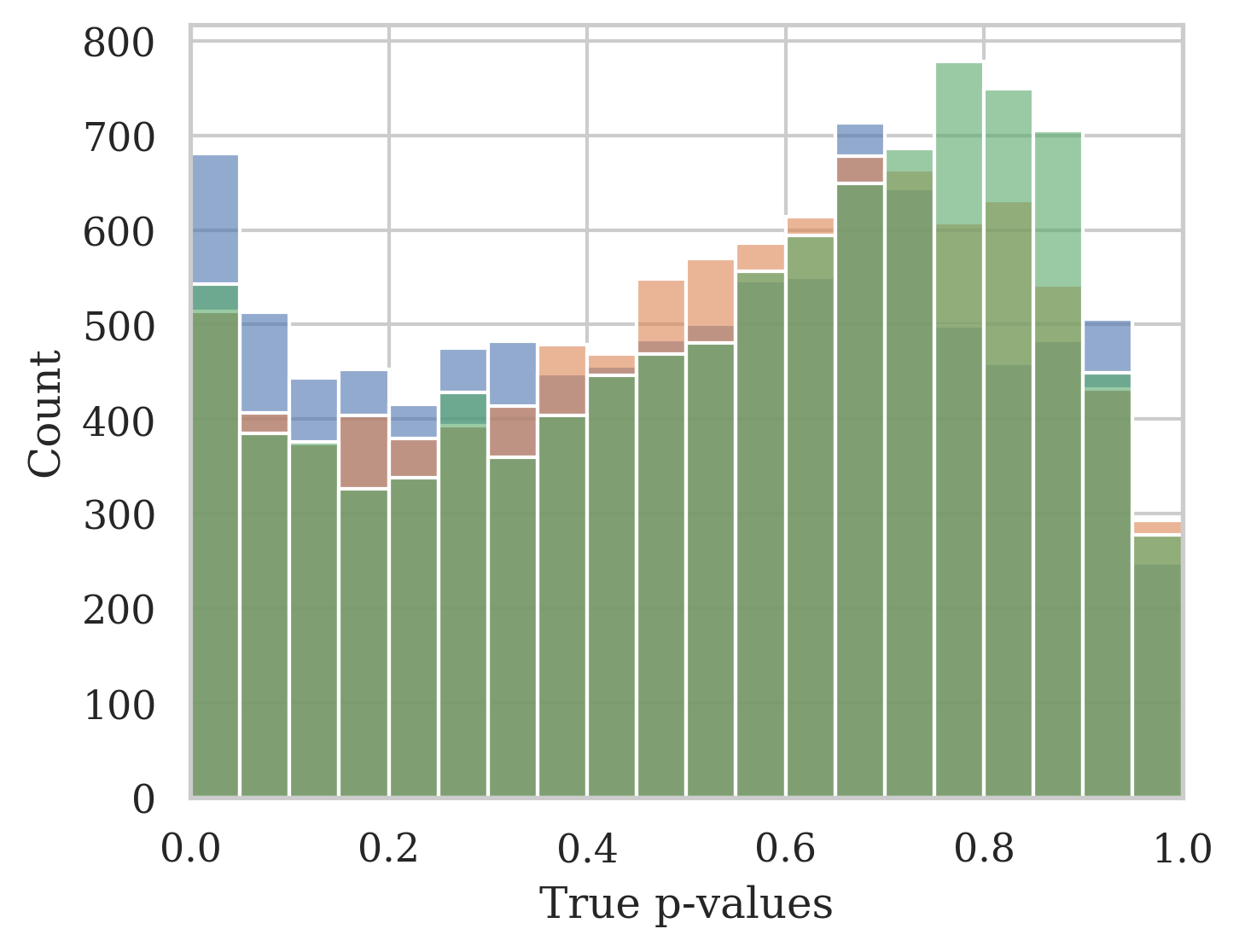}
                    \caption{NN true-class p-values.}
                    \label{fig:nnTruePvalues10}
                \end{subfigure}
                \hfill
                \begin{subfigure}{0.49\textwidth}
                    \centering
                    \includegraphics[width=\textwidth]{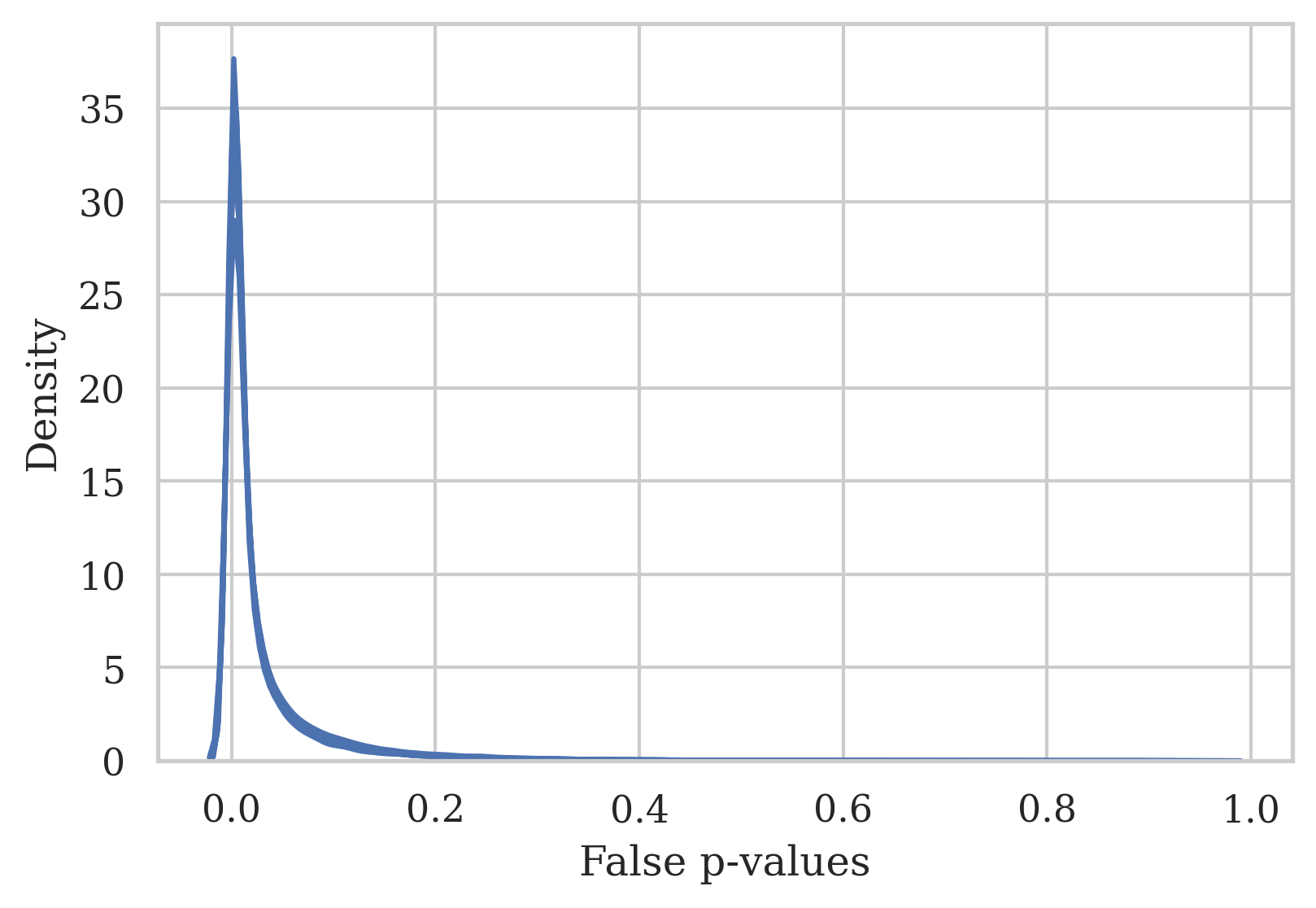}
                    \caption{NN false-class p-values.}
                    \label{fig:nnFalsePvalues10}
                \end{subfigure}
                \caption{10-class classification on MNIST10 with our proposed conformal loss function. The increased training sample count from around 13,000 with binary classification to 60,000 significantly improved our models' calibration and, consequently, their approximate validity significantly.}\label{fig:nn10}
            \end{figure}

        \paragraph{Increasing the number of classes and samples}
        \Cref{fig:nn10} presents our proposed method's performance on the MNIST10 dataset. This involves differentiating between 10 classes (digits 0--9) instead of two (0, 1), and a consequent increase of training samples from around 13,000 to 60,000 (see \Cref{tab:sampleStats}).
        
        The calibration curves became visibly closer to the diagonal (exact validity), smooth, and consistent between model iterations compared to binary classification (\Cref{fig:NnCalibration}). Additionally, all ten models achieve close to exact validity for $0.1 \leq \epsilon \leq 0.3$. The smoothness of the calibration curves was a reflection of the true-class p-value distribution, which more closely followed the uniform distribution $\mathcal{U}_{[0,1]}$.
        
    \subsubsection{Evaluation of all datasets}\label{sec:otherResults}
\begin{figure}[!h]
            \centering
            \begin{subfigure}{0.32\textwidth}
                \centering
                \includegraphics[width=\textwidth]{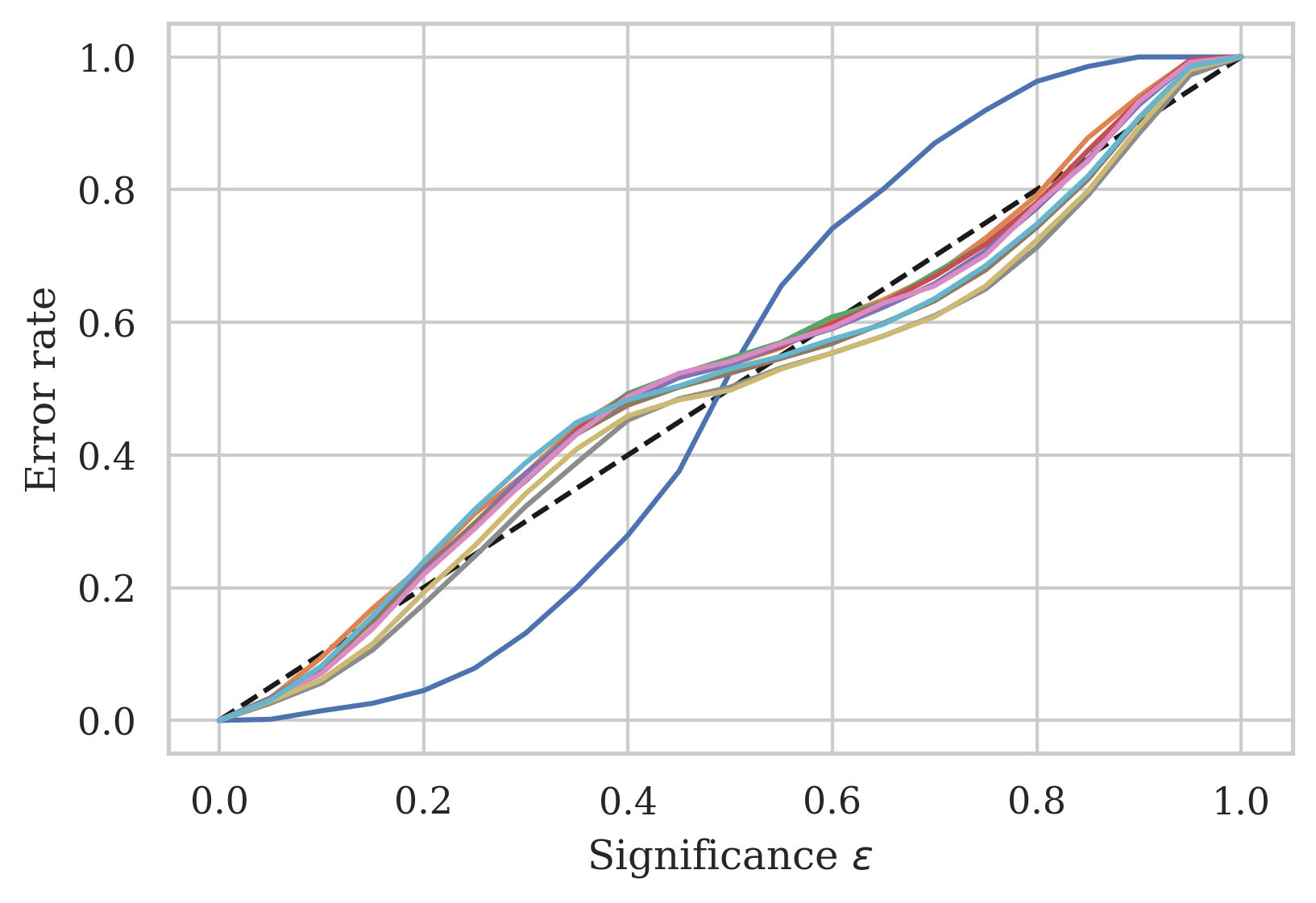}
                \caption{USPS2 dataset.}
                \label{fig:usps2Calibration}
            \end{subfigure}
            \hfill
            \begin{subfigure}{0.32\textwidth}
                \centering
                \includegraphics[width=\textwidth]{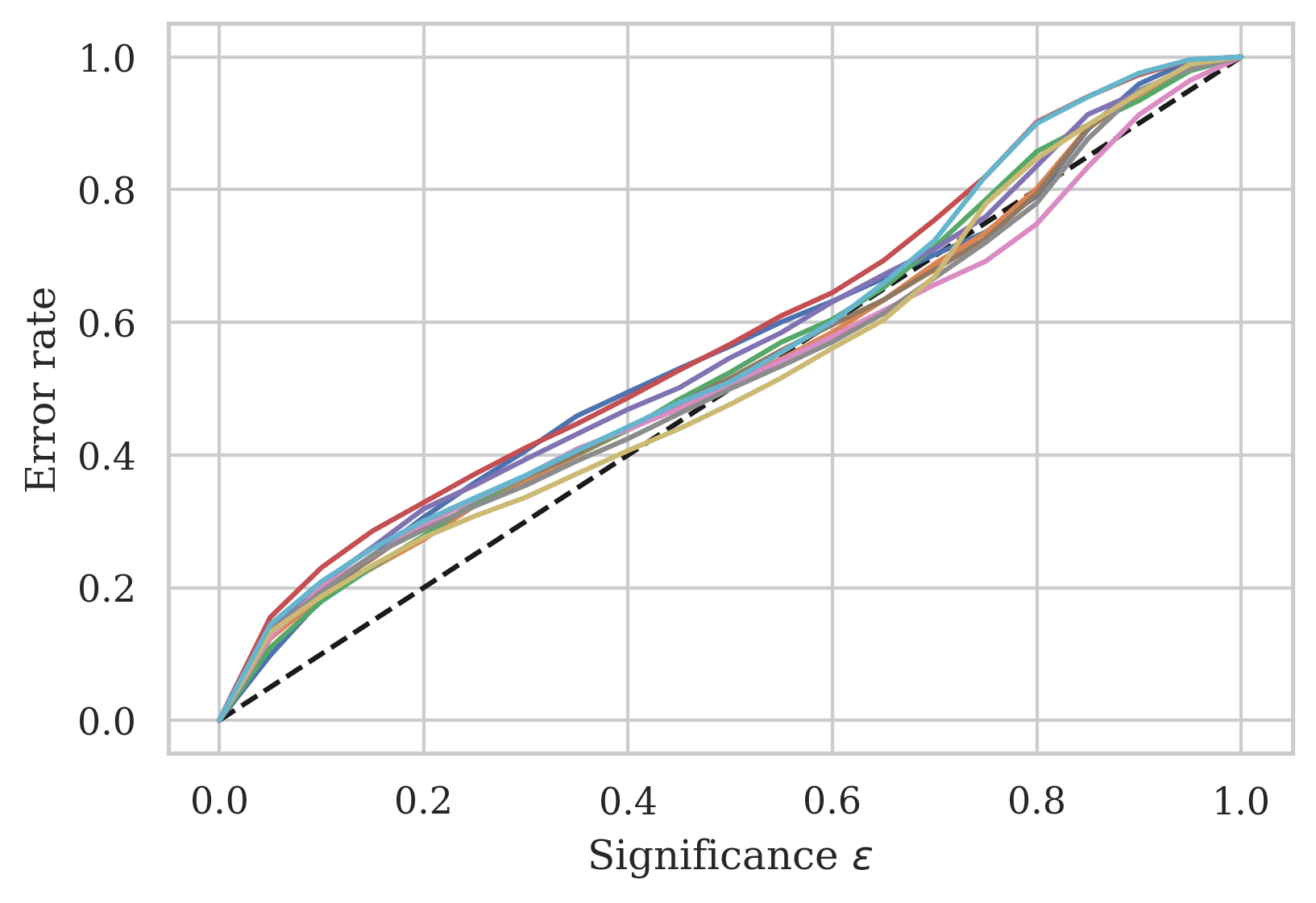}
                \caption{USPS10 dataset.}
                \label{fig:usps10Calibration}
            \end{subfigure}
            \hfill
            \begin{subfigure}{0.32\textwidth}
                \centering
                \includegraphics[width=\textwidth]{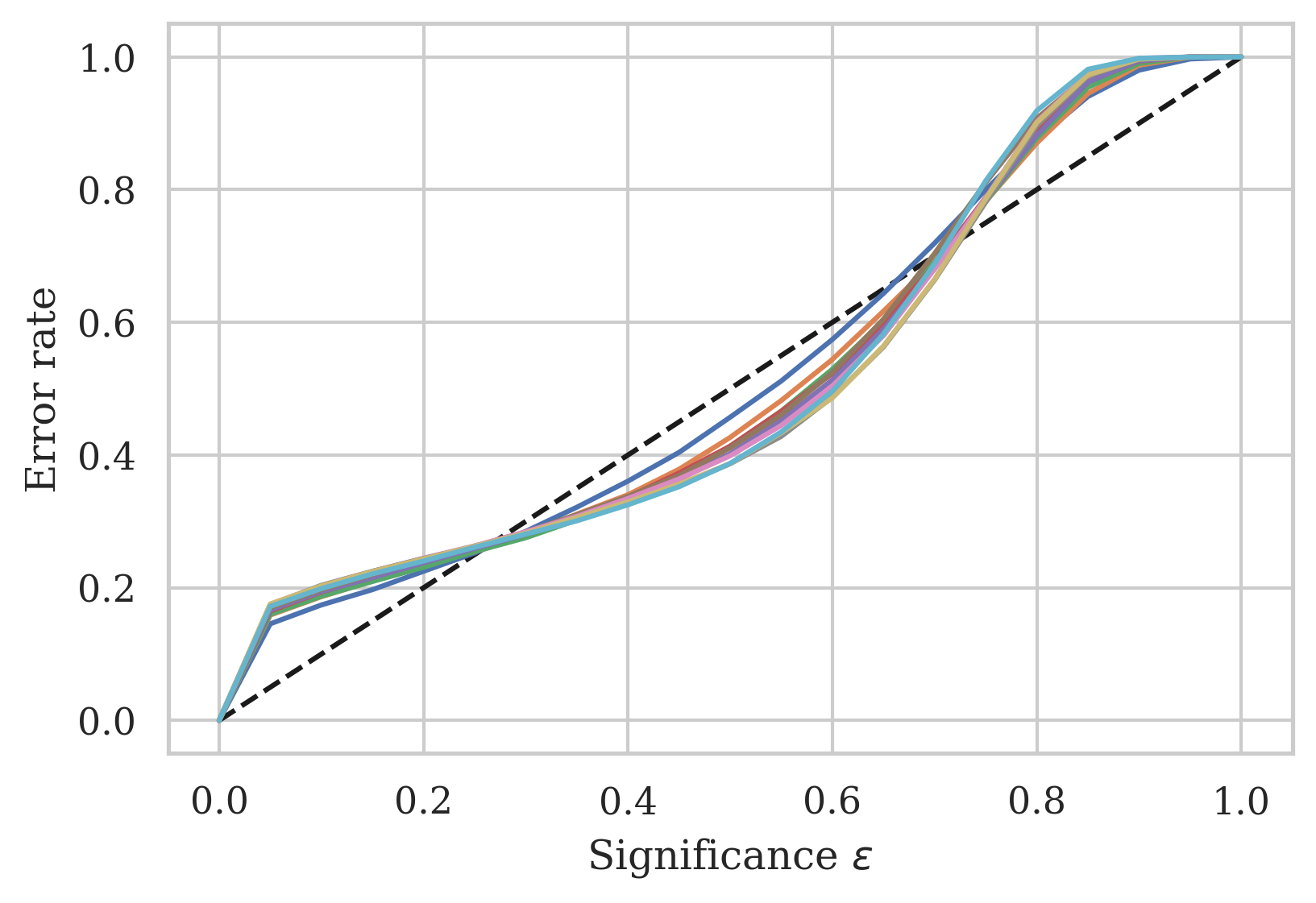}
                \caption{BANK dataset.}
                \label{fig:bankCalibration}
            \end{subfigure}
            \begin{subfigure}{0.32\textwidth}
                \centering
                \includegraphics[width=\textwidth]{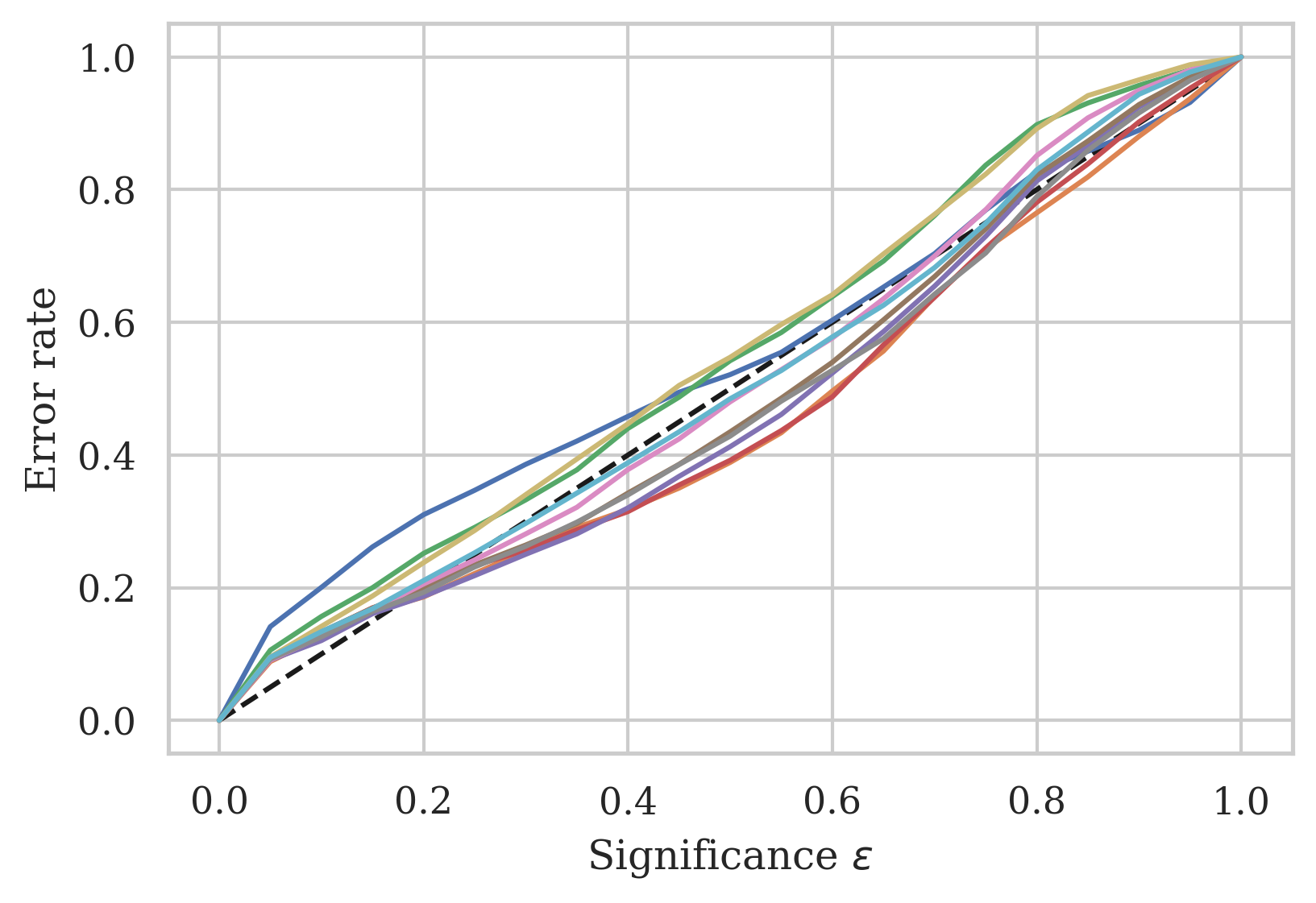}
                \caption{WINE dataset.}
                \label{fig:wineCalibration}
            \end{subfigure}
            \quad
            \begin{subfigure}{0.32\textwidth}
                \centering
                \includegraphics[width=\textwidth]{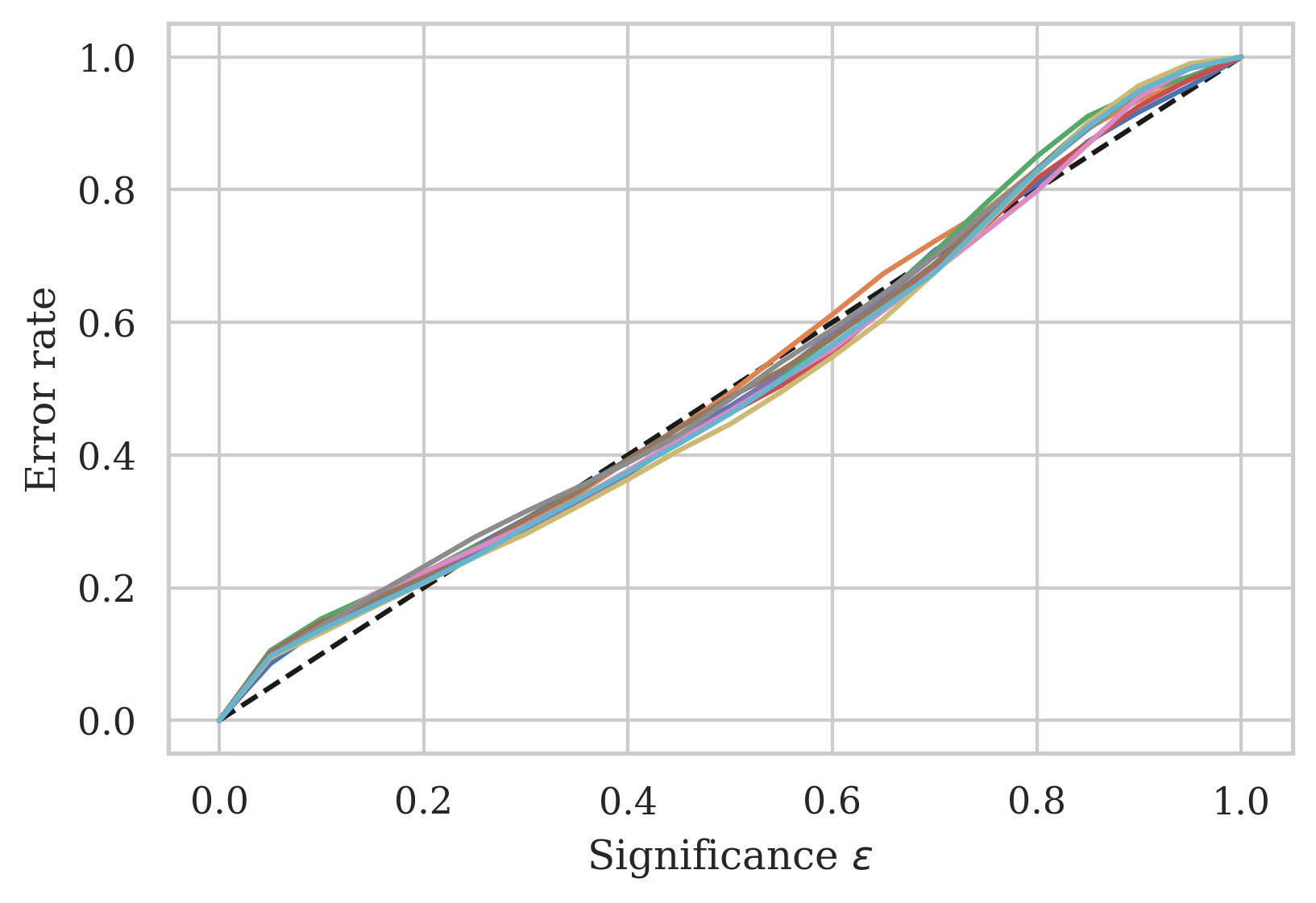}
                \caption{MSHRM dataset.}
                \label{fig:mushroomCalibration}
            \end{subfigure}
            \caption{Calibration lines for 10 iterations each of 5 classification tasks. Using the conformal loss function we derived from the MNIST2 and MNIST10 datasets achieved approximately valid results.}\label{fig:allCalibrationCurves}
        \end{figure}

        \Cref{fig:allCalibrationCurves} shows that the error lines for the remaining datasets (USPS2, USPS10, BANK, WINE, MSHRM) approximately follow the diagonal. This means that our DL approximation is successful, and our conformal loss function can be transferred from the MNIST2 and MNIST10 datasets onto new and unseen datasets with similar performance. The best results are achieved on the USPS10 (multi-class) and MSHRM (binary) datasets, perhaps because of the relatively large sample sizes. However, in comparison, the BANK dataset also has more than 40,000 samples and did not perform as well, so the underlying cause merits further investigation.

        A detailed breakdown of our results and ACP results for comparison is given in \Cref{tab:allDatasetsResults}. Interestingly, NN approximation multi-set prediction rates tend to be significantly lower than ACP rates (e.g., USPS10 with ACP multi-rate $\approx 65\%$ compared to NN multi-rate $= 24\%$). An extreme case may be observed for the WINE dataset (binary), where all predictions are multi-set predictions containing both labels. Furthermore, this trend is accompanied by NN average set sizes being closer to one (optimal prediction) in most cases. This is particularly noticeable for the USPS10 dataset (ACP average $\approx 4.3$). Furthermore, the fuzziness of the NN approximation tends to be slightly improved over ACP results. In other words, the p-values, apart from the largest p-value, tend to be close to 0, which explains the overall low multi-set rates and average set sizes.

        Finally, as expected, the training time is much lower for the NN approach compared to ACP, since only one model must be trained at a time.  

        \setlength{\tabcolsep}{3pt}
        \begin{table}[!h]
        \caption{The performance results of our NN conformal loss approximation approach (average over 10 iterations) compared to ACP with 2, 5, and 10 ensemble models for $\epsilon = 0.1$. Error, Empty, Single, and Multi refer to the prediction set rates. Avg., Miscal., and Fuzz.\ refer to the average set size, miscalibration rate, and p-value fuzziness, respectively. Finally, Time refers to the training time in seconds.}\label{tab:allDatasetsResults}
        \centering\scalebox{0.94}{
        \begin{tabular}{>{}l>{}l>{}r>{}r>{}r>{}r>{}r>{}r>{}r}
            \toprule
            \textbf{Model} & \textbf{Metric} & \textbf{MNIST2} & \textbf{MNIST10} & \textbf{USPS2} & \textbf{USPS10} & \textbf{BANK} & \textbf{WINE} & \textbf{MSHRM} \\
            \specialrule{.4pt}{2pt}{0pt}\midrule
            \multirow{8}{*}{\textbf{NN}} & \textbf{Error} & 5.31\% & 11.43\% & 12.15\% & 19.71\% & 19.39\% & 13.98\% & 14.28\% \\
             & \textbf{Empty} & 5.31\% & 5.34\% & 10.59\% & 12.12\% & 16.35\% & 9.85\% & 13.75\% \\
             & \textbf{Single} & 94.69\% & 48.64\% & 78.22\% & 63.65\% & 83.65\% & 89.85\% & 86.14\% \\
             & \textbf{Multi} & 0.00\% & 46.02\% & 2.70\% & 24.23\% & 0.00\% & 0.30\% & 0.11\% \\
             & \textbf{Avg.} & 0.95 & 1.60 & 0.96 & 1.31 & 0.84 & 0.90 & 0.86 \\
             & \textbf{Miscal.} & 1.11 & 0.73 & 0.90 & 1.01 & 1.26 & 0.80 & 0.48 \\
             & \textbf{Fuzz.} & 0.00 & 0.26 & 0.01 & 0.08 & 0.00 & 0.00 & 0.00 \\
             & \textbf{Time} & 3.64s & 8.29s & 1.48s & 1.18s & 1.87s & 0.78s & 2.53s \\
            \midrule
            \multirow{8}{*}{\textbf{ACP-2}} & \textbf{Error} & 7.28\% & 5.20\% & 2.09\% & 3.64\% & 9.23\% & 9.60\% & 8.19\% \\
             & \textbf{Empty} & 7.28\% & 0.00\% & 1.61\% & 0.00\% & 0.00\% & 0.00\% & 0.00\% \\
             & \textbf{Single} & 92.72\% & 69.30\% & 98.39\% & 28.40\% & 98.10\% & 82.56\% & 98.45\% \\
             & \textbf{Multi} & 0.00\% & 30.70\% & 0.00\% & 71.60\% & 1.90\% & 17.44\% & 1.55\% \\
             & \textbf{Avg.} & 0.93 & 1.89 & 0.98 & 4.68 & 1.02 & 1.17 & 1.02 \\
             & \textbf{Miscal.} & 0.43 & 0.66 & 2.29 & 0.87 & 0.37 & 0.59 & 0.39 \\
             & \textbf{Fuzz.} & 0.00 & 0.30 & 0.00 & 0.90 & 0.02 & 0.05 & 0.02 \\
             & \textbf{Time} & 9.00s & 16.39s & 4.98s & 4.57s & 6.36s & 4.27s & 6.19s \\
            \midrule
            \multirow{8}{*}{\textbf{ACP-5}} & \textbf{Error} & 4.11\% & 4.32\% & 1.28\% & 4.63\% & 8.05\% & 0.00\% & 7.17\% \\
             & \textbf{Empty} & 4.11\% & 0.00\% & 0.00\% & 0.00\% & 0.00\% & 0.00\% & 0.00\% \\
             & \textbf{Single} & 95.89\% & 68.97\% & 88.44\% & 35.28\% & 96.28\% & 0.00\% & 97.45\% \\
             & \textbf{Multi} & 0.00\% & 31.03\% & 11.56\% & 64.72\% & 3.72\% & 100.00\% & 2.55\% \\
             & \textbf{Avg.} & 0.96 & 2.06 & 1.12 & 4.75 & 1.04 & 2.00 & 1.03 \\
             & \textbf{Miscal.} & 2.15 & 0.78 & 3.44 & 0.74 & 0.93 & 3.60 & 0.56 \\
             & \textbf{Fuzz.} & 0.00 & 0.34 & 0.04 & 0.84 & 0.02 & 0.42 & 0.02 \\
             & \textbf{Time} & 18.90s & 43.36s & 11.28s & 11.25s & 15.30s & 11.96s & 19.51s \\
            \midrule
            \multirow{8}{*}{\textbf{ACP-10}} & \textbf{Error} & 4.35\% & 4.67\% & 0.48\% & 2.39\% & 8.97\% & 0.00\% & 6.82\% \\
             & \textbf{Empty} & 4.35\% & 0.00\% & 0.00\% & 0.00\% & 0.00\% & 0.00\% & 0.00\% \\
             & \textbf{Single} & 95.65\% & 81.63\% & 98.56\% & 32.44\% & 98.22\% & 0.00\% & 96.92\% \\
             & \textbf{Multi} & 0.00\% & 18.37\% & 1.44\% & 67.56\% & 1.78\% & 100.00\% & 3.08\% \\
             & \textbf{Avg.} & 0.96 & 1.53 & 1.01 & 3.93 & 1.02 & 2.00 & 1.03 \\
             & \textbf{Miscal.} & 1.68 & 0.93 & 2.21 & 1.14 & 0.85 & 2.71 & 0.61 \\
             & \textbf{Fuzz.} & 0.00 & 0.28 & 0.05 & 0.76 & 0.02 & 0.25 & 0.02 \\
             & \textbf{Time} & 26.96s & 128.01s & 22.16s & 24.35s & 30.21s & 29.18s & 34.84s \\
            \bottomrule
        \end{tabular}}
        \end{table}

\section{Discussion}\label{sec:discussion}
    We presented a comprehensive empirical evaluation of our proposal on 7 classification tasks over 5 benchmark datasets. Results across ten experiment iterations found that our conformal loss function is competitive with ACP for approximate validity and predictive efficiency. The crucial difference is that our direct approach needs only one model and therefore has significantly improved computational efficiency. The computational benefit increases proportionally as the number of ensemble models $n$ in ACP grows.
    
    Our novel loss function minimises the difference between the model output with the expected CP output distribution (\Cref{sec:proposedLoss}). Because the false-class p-values follow the same trend as false-class probability outputs in standard DL classification (values close to 0), our loss component $loss_{false}$ is successful. Consequently, we achieve high predictive efficiency and are comparable with ACP up to $n=10$ ensemble models at low significance levels (\Cref{fig:NnEfficiency}). The optimal and most precise prediction for CP models is $\text{\textbar}\Gamma^{1-\epsilon}\text{\textbar} = 1$ (\Cref{sec:cpBackground}), and our conformal DL models achieve on average $\text{\textbar}\Gamma^{1-0.05}\text{\textbar} = 0.97$, exactly on par with ACP ($n=10$) at $\epsilon=0.05$.
    
    In terms of true-class outputs, we achieve approximate validity on par with ACP for $n>5$ (\Cref{tab:setResults}), measured with the Kolmogorov-Smirnov test for uniformity \cite{zhang2010fast}. ACP tends to be conservative for low significance levels and invalid for higher levels \cite{linusson2017calibration}. In contrast, our proposed method has a calibration line on average closer to the expected diagonal at most individual points.

    Overall, our proposed approach results in `less invalid' predictors compared to ACP models with the same NN underlying algorithm on the 7 classification tasks. The results analysis and interpretation in this article may be specific to the feedforward networks that were used and merits further investigation for further model architectures. However, the loss function derived from the MNIST dataset was successfully transferred to 4 new datasets, which strongly supports the generalisability of our approach.
    
    Our proposed loss function may be refined to train predictors closer to absolute validity by improving the uniformity measure. To maintain differentiability, we measure the deviation through distribution moments (\Cref{sec:proposedLoss}), which are not necessarily unique to the expected distribution $\mathcal{U}_{[0, 1]}$. Empirically, we achieve approximate uniformity and therefore approximate validity, but the distribution measure may be improved in future work. Nonetheless, approximate validity may prove useful for many real-world scenarios.

\section{Conclusion and future directions}
    We propose a novel conformal loss function which approximates the two-step CP framework for classification with one conformal loss function. By learning to output predictions in the distribution characteristic to CP from the input data (\Cref{sec:CP}), models trained with our loss function skip the intermediate non-conformity score, hence, reducing the inherent algorithmic complexity. The loss function is fully differentiable and compatible with any gradient descent-based Deep Learning neural network (\Cref{sec:proposedMethod}). Our novel approach is most successful for small significance levels, which in practice are of high interest to guarantee low error rates. Additionally, our direct conformal p-value prediction has the potential to simplify CP optimisation with further study.
    
    We carried out an extensive and rigorous empirical evaluation of our proposed method for 7 classification tasks on 5 benchmark datasets, with three main insights:
    \begin{itemize}
        \item Our proposal is comparative to Aggregated Conformal Prediction (ACP) for low significance levels, a common CP variant that has been successful on real-world datasets with high predictive efficiency (see \Cref{sec:cpBackground}). We achieve results on par with ACP for both approximate validity and predictive efficiency when $n \geq 3$ ensemble models.
        \item Notably, our conformal loss function significantly improves ACP's computational efficiency without compromising CP performance. We train only one model, compared to $n$ ACP ensemble models. As $n$ increases, the computational savings of our method grow proportionally.
        \item Finally, our one-step CP approximation reduces the algorithmic complexity of the traditionally two-step CP framework. This provides a potential new avenue for CP optimisation research, which is traditionally a difficult task since the interaction between the underlying algorithm and the non-conformity measure have to be considered simultaneously (see \Cref{sec:relatedWork}).
    \end{itemize}

    Future work includes improving our conformal loss function to achieve guaranteed validity over approximate validity. For example, this may be achieved by improving the output uniformity metric to be more precise (\Cref{sec:discussion}). Additionally, our empirical study may be extended with a theoretical evaluation and optimisation of the loss function for gradient descent to improve model convergence.


\section*{Declarations}
\begin{itemize}
\item Funding: This research is funded by University of Brighton’s `Rising Stars' research grant, and Innovate UK's AKT2I grant `Machine vision segmentation for automated UK train tracking and railway maintenance'.
\item Conflict of interest/Competing interests: The authors declare that they have no conflict of interest.
\item {Availability of data and materials: The 5 datasets analysed during the study are publicly available as follows: The MNIST dataset~\cite{lecun1998gradient} at \url{http://yann.lecun.com/exdb/mnist/}; The USPS dataset~\cite{hull1994database} at \url{https://www.kaggle.com/datasets/bistaumanga/usps-dataset}; And the WINE~\cite{CorCer09}, BANK~\cite{moro2014data}, and MSHRM~\cite{schlimmer1981mushroom} datasets were obtained from the UCI Machine Learning Repository~\cite{Dua2019} at \url{https://archive.ics.uci.edu}.}
\item Code availability: The code is available at \url{https://github.com/juliameister/dl-confident-loss-function}.
\end{itemize}








\bibliography{sn-bibliography}


\end{document}